%% file: arXiv-ml-robustness.tex
\documentclass{egpubl}

 \JournalSubmission
      
\usepackage[T1]{fontenc}
\usepackage{dfadobe}  

\usepackage{cite}  

\BibtexOrBiblatex

\electronicVersion
\PrintedOrElectronic

\ifpdf \usepackage[pdftex]{graphicx} \pdfcompresslevel=9
\else \usepackage[dvips]{graphicx} \fi

\usepackage{egweblnk}

\graphicspath{{./figs/}} 

\usepackage{amssymb}
\usepackage{amsmath} 
\usepackage{amsfonts}
\usepackage{algorithm}
\usepackage{algpseudocode}
\usepackage{hyperref}
\usepackage[table]{xcolor}
\usepackage{tabu} 
\usepackage{pbox}
\newcommand{\para}[1]        {\vspace{0pt}\noindent{\textbf{#1}}}

\newtheorem{lemma}{Lemma}

\newcommand {\mm}[1] {\ifmmode{#1}\else{\mbox{\(#1\)}}\fi}

\newcommand{\Xspace}        {\mm{\mathbb{X}}}

\newcommand{\Rspace}        {\mm{\mathbb{R}}}

\newcommand{\minR}        {\mm{\mathrm{minR}}}

\newcommand{\mydeg}{\mm{\mathrm{deg}}}

\newcommand{\etal}{{et al.}}
\newcommand{\eg}{{e.g.}}
\newcommand{\ie}{{i.e.}}
\newcommand{\wrt}{{w.r.t.}}

\newcommand{\EW}        {\mm{\mathsf{E3SM\,Wind}}}
\newcommand{\EWL}        {\mm{\mathsf{E3SM\,Wind\,L}}}
\newcommand{\HK}        {\mm{\mathsf{Hurrican\,Katrina}}}
\newcommand{\ME}        {\mm{\mathsf{MPAS\,O\,Eddy}}}

\newcommand{\myupdate}[1]{{\color{blue}{#1}}}

\title[Multilevel Robustness]
      {Multilevel Robustness for 2D Vector Field Feature Tracking, Selection, and Comparison}
\author[L. Yan, P. Ullrich, L. Roekel, B. Wang, H. Guo]
{\parbox{\textwidth}{\centering Lin Yan$^{1}$, Paul Aaron Ullrich$^{2}$, Luke P. Van Roekel$^{3}$, Bei Wang$^{4}$,  Hanqi Guo$^{5}$}
\\ 
{\parbox{\textwidth}{\centering $^1$ Environmental Science \& Mathematics and Computer Science, Argonne National Laboratory, USA\\
$^2$ Department of Land, Air and Water Resources, University of California, Davis, USA\\
$^3$ The Fluid Dynamics and Solid Mechanics Group (T-3), Los Alamos National Laboratory, USA\\
$^4$School of Computing, Scientific Computing and Imaging (SCI) Institute, University of Utah, USA\\
$^5$ Department of Computer Science and Engineering, The Ohio State University, USA
}}}

\begin{document}

\maketitle
\begin{abstract}
\input{sec-abstract.tex}
\end{abstract}  

\input{sec-introduction}

\input{sec-related-work}

\input{sec-background}

\input{sec-ml-robustness}
\input{sec-visualization}

\input{sec-results}

\input{sec-discussions}

\section*{Acknowledgments} 
This material is based upon work supported by the U.S. Department of Energy (DOE), Office of Science, under contract number DE-AC02-06CH11357. This work is also supported by the U.S. Department of Energy, Office of Advanced Scientific Computing Research, Scientific Discovery through Advanced Computing (SciDAC) program. This work is also in part supported by NSF IIS-1910733 and DOE DE-SC0021015.
\input{ml-robustness-refs.bbl}

\end{document}

%% file: sec-abstract.tex
Critical point tracking is a core topic in scientific visualization for understanding the dynamic behavior of time-varying vector field data. 
The topological notion of robustness has been introduced recently to quantify the structural stability of critical points, that is, the robustness of a critical point is the minimum amount of perturbation to the vector field necessary to cancel it. 
A theoretical basis has been established previously that relates critical point tracking with the notion of robustness, in particular, critical points could be tracked based on their closeness in stability, measured by robustness, instead of just distance proximities within the domain.  
However, in practice, the computation of classic robustness may produce artifacts when a critical point is close to the boundary of the domain; thus, we do not have a complete picture of the vector field behavior within its local neighborhood. 
To alleviate these issues, we introduce a multilevel robustness framework for the study of 2D time-varying vector fields. 
We compute the robustness of critical points across varying neighborhoods to capture the multiscale nature of the data and to mitigate the boundary effect suffered by the classic robustness computation. 
We demonstrate via experiments that such a new notion of robustness can be combined seamlessly with existing feature tracking algorithms to improve the visual interpretability of vector fields in terms of feature tracking, selection, and comparison for large-scale scientific simulations. 
We observe, for the first time, that the minimum multilevel robustness is highly correlated with physical quantities used by domain scientists in studying a real-world tropical cyclone dataset. 
Such an observation helps to increase the physical interpretability of robustness.

%% file: sec-introduction.tex
\section{Introduction}
\label{sec:introduction}

The analysis and visualization of vector fields has seen widespread applications in science and engineering, including combustion, climate study, and ocean modeling. 
With the increasing size and complexity of vector field data that arise from scientific simulations, vector field topology has been one of the most promising tools to describe and interpret vector field behavior by providing meaningful abstraction and summarization~\cite{PobitzerPeikertFuchs2011, BujackYanHotz2020}. 

Critical points (\ie, where a vector field vanishes) are core features of vector field topology. 
To improve the visual interpretability of time-varying vector fields, a key challenge is feature tracking~\cite{PostVrolijkHauser2003} -- in particular, critical point tracking -- that is, to resolve the correspondences between critical points in successive time steps in the form of trajectories, and to understand the dynamic behavior of these trajectories via selections and comparisons. 

The topological notion of robustness has been introduced recently to quantify the stability of critical points. 
The robustness of a critical point is defined to be the minimum amount of perturbation to the vector field necessary to cancel it. 
Robustness has been shown to be useful in feature extraction~\cite{WangBujackRosen2017} and simplification~\cite{SkrabaWangChen2014,SkrabaWangChen2015,SkrabaRosenWang2016} of vector field data.  
In particular, Skraba and Wang inferred  correspondences between critical points based on their closeness in stability, measured by robustness, instead of just distance proximities within the domain~\cite{SkrabaWang2014b}. 
They obtained theoretical results by relating critical point tracking with the notion of robustness: roughly speaking, critical points with high robustness values could be tracked more easily and more accurately~\cite{SkrabaWang2014b}. 
However, the results in~\cite{SkrabaWang2014b} were theoretical in nature, and bringing this theory to practice is nontrivial. 
Vector field data generated from large-scale ocean, atmospheric, and fluid dynamics simulations contain features at different scales. 
It is a common practice for researchers to study the data within a chosen domain of interest. 
For critical points close to the boundary of the domain, we have an incomplete picture of flow behavior within their local neighborhoods. 
Consequently, the computation of classic robustness may suffer from poor boundary conditions; for instance, a critical point may not find a cancellation partner or may be forced to cancel with another critical point that is far away in the known data domain (see~\autoref{sec:ml-robustness} for details). 
Such phenomena decrease the effectiveness in robustness-based critical point tracking. 

In this paper, we introduce \emph{multilevel robustness} for critical points, a ``scale-aware'' notion of robustness that accommodates the inherent multiscale nature of vector field data. 
Multilevel robustness helps to mitigate the boundary effect suffered by the classic robustness computation.  
More importantly, it can be integrated with existing feature tracking algorithms to improve feature tracking, selection, and comparison.

\para{Contributions.} 
Building upon the theoretical basis established previously~\cite{SkrabaWang2014b}, the focus of this paper is to realize robustness-based critical point tracking in practice for large-scale scientific simulations. To that end,  
\begin{itemize}
\item We introduce a multilevel robustness framework for the study of 2D time-varying vector fields. We compute the robustness of critical points across varying neighborhoods to capture the multiscale nature of the data and to mitigate the boundary effect suffered by the classic robustness computation. 
\item We demonstrate that our proposed framework -- in particular, the minimum multilevel robustness -- can be combined with feature tracking algorithms such as FTK~\cite{GuoLenzXu2021} to improve the visual interpretability of vector fields in terms of feature tracking, selection, and comparison.
\item We observe, \emph{for the first time}, that the minimum multilevel robustness is highly correlated with physical quantities (such as maximum wind speed and mean sea-level pressure) used by domain scientists in studying a real-world tropical cyclone dataset.   
\end{itemize}
The observation above is quite exciting as it  implies that robustness -- a notion of feature stability derived based on vector field perturbation -- is highly correlated with scalar-valued physical quantities commonly used by domain scientists to study tropical cyclones, which helps to increase the physical interpretability of robustness.

%% file: sec-related-work.tex
\section{Related Work}
\label{sec:related-work}

We review related work on vector field topology, critical point tracking, and robustness of critical points. 

\para{Vector field topology} 
has been researched over the past decades since it was firstly introduced by Helman and Hesselink~\cite{HelmanHesselink1989}. 
However, as pointed out by Pobitzer~\etal~\cite{PobitzerPeikertFuchs2011} and Bujack~\etal~\cite{BujackYanHotz2020}, vector field topology for time-varying flows remains a challenge. In particular, it is difficult to
interpret flow topology {\wrt} physical meaning in the time-varying setting~\cite{BujackYanHotz2020}. 
In this paper, we focus on the tracking and visualization of critical points of time-varying vector fields, and investigate the potential relationship between the topological properties of critical points and physical quantities of relevance to real-world flow dataset. 

\para{Critical point tracking}, which reconstructs the trajectories of critical points over time, may be achieved by proximity-, integral-, and interpolation-based methods. 
Proximity-based critical point tracking includes the work of  Helman and Hesselink~\cite{HelmanHesselink1989, HelmanHesselink1990}, which connects the critical points (singularities) from separate time steps based on proximity and region connectedness. 

For integral-based critical point tracking approaches, Theisel and Seidel~\cite{TheiselSeidel2003} recast the tracking of critical points in a 2D vector field as an integration problem in a 3D field, called feature flow field (FFF), and computed feature trajectories based on tangent curves in FFF. 
Weinkauf~\etal~\cite{WeinkaufTheiselVan2010} improved upon the FFF and presented a more stable formulation for tracking critical points by addressing instabilities in the numerical integration during the computation of tangent curves. This is followed by the work in~\cite{ReininghausKastenWeinkauf2011} that introduced a combinatorial version of FFF.

An example of interpolation-based method is from Tricoche~\etal~\cite{TricocheScheuermannHagen2001a, TricocheWischgollScheuermann2002}, who implemented the linear interpolation between time steps, which  guarantees the existence of one critical point in each cell, and analyzed the cell faces to detect changes in the topology over time. 
Analogously to~\cite{TricocheScheuermannHagen2001a}, Garth~\etal~\cite{GarthTricocheScheuermann2004} extended this approach and provided a critical point tracking algorithm for 3D time-varying vector fields. 
Guo~\etal~\cite{GuoLenzXu2021} proposed a \emph{simplicial} spacetime meshing scheme for tracking critical points, referred to as the Feature Tracking Kit (FTK) framework, which is further reviewed in \autoref{sec:background}. 

\para{Robustness of critical points} has been introduced recently to quantify the structural stability of critical points with respect to perturbations to the vector fields~\cite{SkrabaWangChen2014,SkrabaWangChen2015,SkrabaRosenWang2016}.   
Robustness has been shown to be useful for the analysis and visualization of vector fields.  For example, Wang~\etal~\cite{WangRosenPrimoz2013} studied how the robustness of a critical point evolves in the time-varying setting.  Skraba and Wang~\cite{SkrabaWang2014b} showed potential usage of robustness in feature tracking, that is, critical points with high robustness values could be tracked more easily and more accurately. 
Robustness is also used 
for 2D~\cite{SkrabaWangChen2014,SkrabaWangChen2015} and 3D~\cite{SkrabaRosenWang2016} vector field simplification.
Lately, Wang~\etal~\cite{WangBujackRosen2017} further extended the classic definition of robustness to a Galilean invariant robustness framework that quantifies the stability of critical points across different frames of reference.
The notion of robustness was further extended to study the stability of degenerate points in tensor fields~\cite{WangHotz2017,JankowaiWangHotz2019}.

The concept of robustness, first introduced by Edelsbrunner~\etal~\cite{EdelsbrunnerMorozovPatel2011a, EdelsbrunnerMorozovPatel2011b}, is closely related to the notion of \emph{persistence}~\cite{EdelsbrunnerLetscherZomorodian2002} -- a common tool used to quantify feature importance. 
In addition to robustness, other measures have been explored to characterize the importance of vector field critical points based on their lifetime~\cite{KastenHotzNoack2011} and scales~\cite{KleinErtl2007}. 

Different to previous efforts, this paper introduces a new notation of \emph{multilevel robustness} for critical points. 
Multilevel robustness studies the robustness of a critical point {\wrt} their local neighborhoods of varying sizes, and thus helps to mitigate the boundary effects suffered by classic robustness computation, and better differentiates the behaviors of  critical points across multiple scales.

%% file: sec-background.tex
\section{Technical Background}
\label{sec:background}

We review the classic notion of robustness and the critical point tracking method by Guo \etal~\cite{GuoLenzXu2021}, referred to as the FTK algorithm in this paper.  

\subsection{Robustness}
\label{sec:robustness}

\begin{figure}[!t]
    \centering
    \includegraphics[width=0.98\columnwidth]{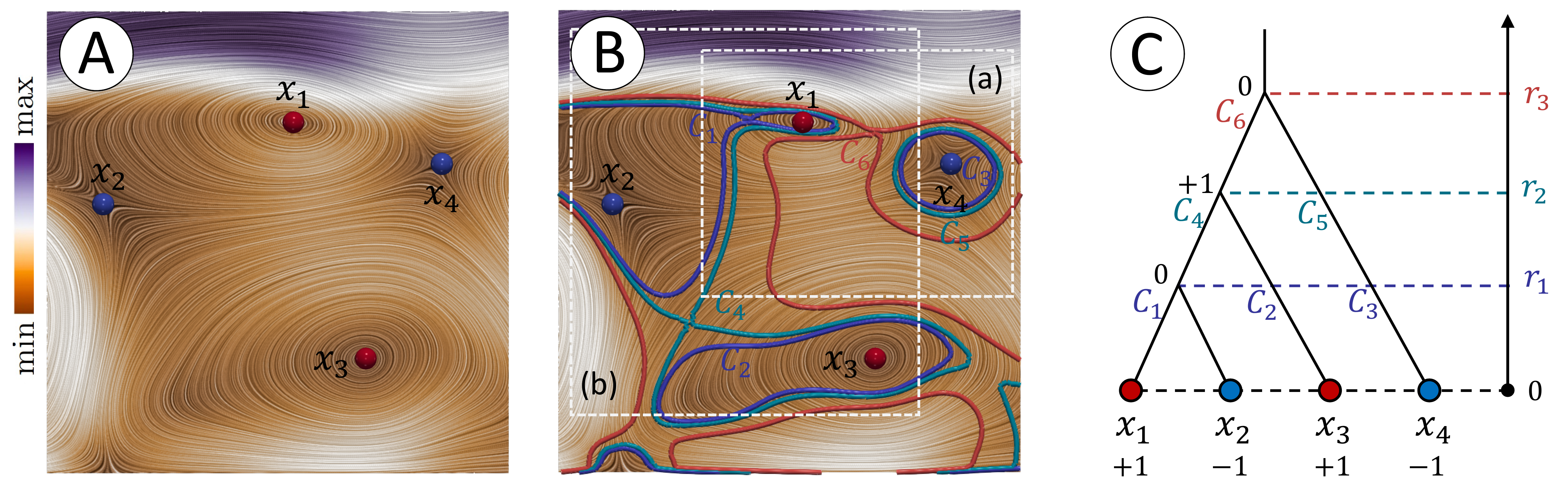}
    \vspace{-2mm}
    \caption{An augmented merge tree generated from a 2D vector field: (A) a continuous 2D vector field $f$; (B)  relations among connected components of sublevel sets $\Xspace_r$; and (C) the augmented merge tree. Sources/sinks/centers are in red, saddles in blue.} 
    \label{fig:example-mt}
    \vspace{-2mm}
\end{figure}

\para{Degrees of critical points.} 
Consider a continuous vector field $f: \Xspace \subseteq \Rspace^2 \to \Rspace^2$ defined on a 2D domain $\Xspace$. A \emph{critical point} $x \in \Xspace$ is an isolated zero in the vector field, that is, $|f(x)| = 0$. 
A critical point $x$ in 2D can be classified with respect to its \emph{degree}, denoted as $\mydeg(x)$, as the number of field rotations while traveling along a closed curve counterclockwise surrounding $x$ enclosing no other critical point. 
In 2D, a saddle point has degree $-1$, whereas a source/sink/center has degree $+1$. 
A connected component $C \subseteq \Xspace$ that contains $n$ critical points $\{x_1, x_2, \cdots, x_n\}$ has a degree that is the sum of the degrees of $x_i$, $\mydeg(C)=\sum_{i=1}^n\mydeg(x_i)$; see~\cite[page 134]{Hatcher2002} for a formal investigation of the degree of a continuous mapping. As illustrated in a 2D vector field in~\autoref{fig:example-mt}(A), $x_1$ and $x_3$ are centers with $+1$ degree, and $x_3$ and $x_4$ are saddles with $-1$ degree.   

\para{Merge tree.} 
The computation of robustness relies on the notion of an augmented merge tree. 
Given a continuous 2D vector field $f:\Xspace \to \Rspace^2$, we can define a scalar field $f_0:\Xspace \to \Rspace$ that assigns the vector magnitude to each point $x \in \Xspace$, that is, $f_0(x) = ||f(x)||_2$. 
Let $\Xspace_r=f_0^{-1} (-\infty, r]$ denote the \emph{sublevel set} of $f_0$ for some $r \geq 0$. $\Xspace_0$ is precisely the set of critical points of $f$. 
In~\autoref{fig:example-mt}(B), $f_0$ is visualized using an orange to purple colormap, and certain sublevel sets $\Xspace_r$ are shown as colored curves.    

We can construct a merge tree of $f_0$ that tracks the evolution of connected components in $\Xspace_r$ as $r$ increases. 
Specifically, leaves in a merge tree represent the creation of a component at a local minimum of $f_0$, internal nodes represent the merging of components, and the root represents the entire space as a single component; see~\autoref{fig:example-mt}(C) for an example.  Once the merge tree is constructed, it can be further augmented with the degrees of critical points (on leaves), and the degrees of components (on internal nodes). 
As shown in~\autoref{fig:example-mt}(C), we use $C_i$ to represent the connected components of the sublevel sets of $\Xspace_r$ for some $r$, and augment the corresponding merge tree with $\deg(x_i)$ and $\deg(C_i)$ as attributes of the tree nodes. 
For example, $C_1$ is one of the three components of $\Xspace_{r_1}$, which contains critical points $x_1$ and $x_2$. 
Therefore, we have $\mydeg(C_1)=\mydeg(x_1)+\mydeg(x_2)=0$.

\para{Robustness.}
The \emph{robustness} of a critical point is the function value of its lowest zero degree ancestor in the merge tree~\cite{WangRosenPrimoz2013}. 
For the example in~\autoref{fig:example-mt}, the robustness of $x_1$ and $x_2$ is $r_1$ and the robustness of $x_3$ and $x_4$ is $r_3$, respectively. 
We review some properties of robustness here for completeness; see~\cite{WangRosenPrimoz2013} for details. 

Let us first define the concept of vector field \emph{perturbation}. A continuous mapping $h: \Xspace \to \Rspace^2$ is an \emph{$r$-perturbation} of $f$, if $d(f, h) \leq r$, where $d(f, h)=\sup_{x\in \Xspace}||f(x)-h(x)||_2$, where $\sup$ means supremum. Suppose a critical point $x$ of $f$ has robustness $r$, then we have:

\begin{lemma}[Critical Point Cancellation\cite{WangRosenPrimoz2013}]
\label{lemma:cancel}
Let $C$ be the connected component of $\Xspace_{r+\eta}$ containing $x$, for an arbitrarily small $\eta > 0$. Then, there exists an $(r+\eta)$-perturbation $h$ of $f$, such that $h^{-1}(0)\bigcap C=\emptyset$ and $h=f$ except possibly within the interior of $C$.
\end{lemma}

\begin{lemma}[Degree Preservation~\cite{WangRosenPrimoz2013}]
\label{lemma:degree}
Let $C$ be the connected component of $\Xspace_{r-\eta}$ containing $x$, for some $0<\eta < r$. For any $\varepsilon$-perturbation $h$ of $f$, where $\varepsilon<r-\eta$, $\mydeg(h^{-1}(0)\bigcap C)= \mydeg(C)$. If $C$ contains only one critical point $x$, $\mydeg(h^{-1}(0)\bigcap C)=\mydeg(x)$.
\end{lemma}

These two lemmas imply that the topological notion of robustness quantifies the stability of a critical point with respect to perturbations of the vector fields. 
Intuitively, Lemma~\autoref{lemma:cancel} implies that a critical point $x$ with a robustness of $r$ may be canceled with a $(r+\eta)$-perturbation, for arbitrarily small $\eta>0$. 
Lemma~\autoref{lemma:degree} states that $x$ may not be canceled with a $(r-\eta)$-perturbation.

\para{Limitations in computing the classic robustness.}
In practice, the robustness of a critical point depends on its cancellation partner(s) defined by the merge tree, whose locations may be influenced by the boundary condition of the known data domain.  
To compute the classic notion of robustness, we use the known data domain to construct a single merge tree, as shown in~\autoref{fig:example-mt}(C). 
If the domain is without boundary, we expect all critical points to have cancellation partners and all the robustness values to be finite (however, there is a technical condition on the domain for the algorithm to work, \ie, trivial tangent bundle, which excludes the sphere). 
If the domain has a boundary, a critical point may be canceled with a potentially far away critical point,  based on the merge tree construction. 
For example, $x_1$ has a partner $x_2$ in~\autoref{fig:example-mt}(B); however, in the cropped region (a), $x_1$ has a new partner $x_4$ since $x_4$ is the only candidate in (a) that may be canceled with $x_1$. 
Furthermore, a critical point may have an infinite robustness value if it does not have a cancellation partner in the known data domain. 
For example, $x_3$ has a partner in the original domain of ~\autoref{fig:example-mt}(B); however, it loses its partner in the cropped region (b). 
These cases happen when the sublevel sets intersect the boundary of the domain where we have an incomplete picture of the flow behavior closer to the boundary.  
We aim to mitigate some of these boundary effects by introducing the notion of multilevel robustness (see~\autoref{sec:ml-robustness}). 

\subsection{Critical Point Trajectories}
Critical point tracking algorithms take a time-varying vector field as the input, and produce as the output 1D geometries that represent the trajectories of critical points in spacetime.  
In general, our multilevel robustness framework may be used to enhance any critical point tracking result; we choose to use the recent FTK algorithm by Guo \etal~\cite{GuoLenzXu2021} for its simplicity and performance.  


\begin{figure*}
    \centering
    \includegraphics[width=2.1\columnwidth]{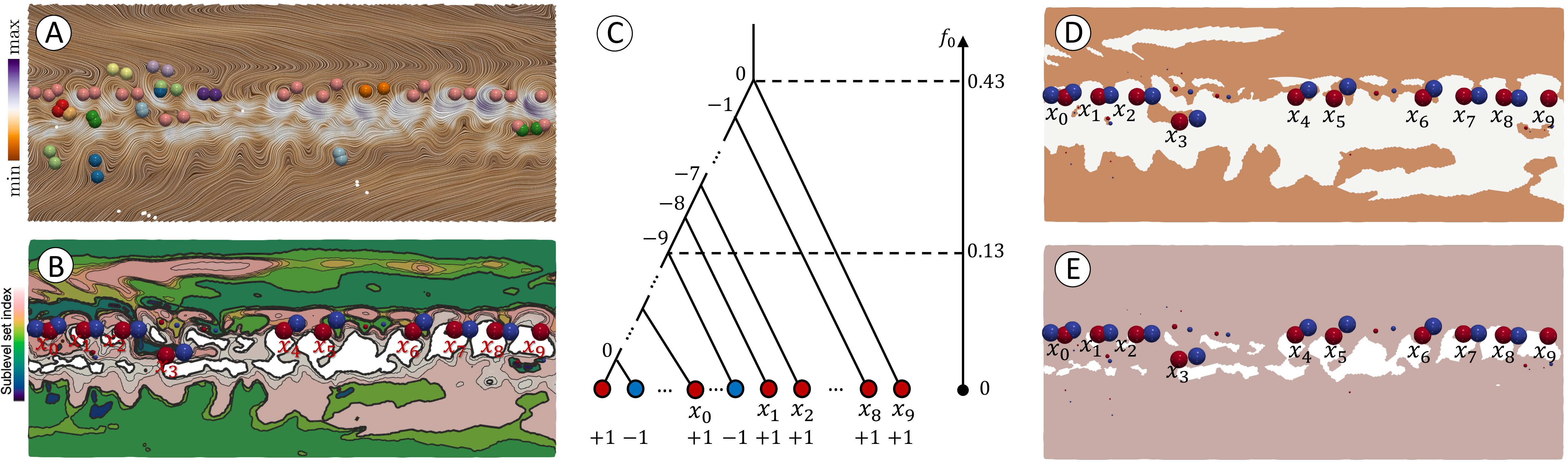}
    \vspace{-4mm}
\caption{Classic robustness analysis of an instance from the $\ME$ dataset. 
(A) A 2D vector field $f$ where critical point partners share the same color. 
(B) Multiple sublevel sets of $f_0$, where critical points are colored by their degree ($+1$ in blue and $-1$ in red), and the size of each critical points is shown to be proportional to their robustness values.   
(C) Part of an augmented merge tree of $f_0$ used in the robustness computation.
(D) Sublevel set $\Xspace_{0.13}$ in orange. 
(E) Sublevel set $\Xspace_{0.43}$ in pink.}
    \label{fig:classic-robustness}
    \vspace{-2mm}
\end{figure*}

\para{Trajectories.} 
Let $\hat{f}: \hat{\Xspace} = \Xspace \times \Rspace \to \Rspace^2$ denote a time-varying vector field over a 2D domain $\Xspace$, where $\hat{f}_t(x) = \hat{f}(x,t): \Xspace \to \Rspace^2$ represents a 2D vector field at time $t \in \Rspace$.  
We define \emph{critical point trajectories} (or simply \emph{trajectories}) as the $0$-levelset of $\hat{f}$, $\hat{\Xspace}_0 := \hat{f}^{-1}(0,0)$, that is,  the vicinity where both $x$- and $y$- components of $\hat{f}$ are $0$ and thus is the intersection of two isosurfaces of both vector components. 

\para{Piecewise linear assumption.} The basic assumption of the tracking method is that $\hat{f}$ is \emph{piecewise linear in spacetime}.  That is, $\hat{\Xspace}$ is a 3D simplicial complex consisting of a set of spacetime tetrahedra $\{T_i\}$ such that 
  $\hat{f}(x) = a_i x + b_i, x\in T_i\subset \hat{\Xspace}$,
where $a_i$ and $b_i$ are constants for each tetrahedron $T_i$, and $\hat{f}$ is $0$-continuous on cell boundaries.  
If the linear system $\hat{f}=0$ in $T_i$ is nondegenerate, the $0$-levelset of $\hat{f}$ in $T_i$ may be analytically solved as a linear curve; otherwise,  degenerate cases may be handled with the simulation of simplicity~\cite{EdelsbrunnerMucke1990}. 
Therefore, trajectories can be extracted as 1D piecewise linear curves in 3D spacetime; see~\cite{GuoLenzXu2021} for details on the construction of spacetime simplicial complexes, handling of degenerates, and extraction of trajectories. 

\para{Interpretation of trajectories.} Note that the $0$-levelset of $\hat{f}$ is not a bijection of time $t$ onto the trajectory; one may observe non-monotonous time along the same trajectory, such as a loop.  The change of monotonicity typically indicates a bifurcation (split) or annihilation (merge). 
Such events may reflect topological changes of the vector field or are simply caused by numerical instabilities in trajectory extraction.  One may need to simplify, segment, and filter the trajectories to understand the vector field dynamics.  To these ends, we demonstrate novel understanding of trajectories based on multilevel robustness, as demonstrated in the rest of this paper.

%% file: sec-ml-robustness.tex
\section{Our New Definition: Multilevel Robustness}
\label{sec:ml-robustness}
 
To mitigate the drawbacks of the classic robustness computation, we introduce a multilevel robustness framework. 
In \autoref{fig:classic-robustness}, we give an example of a classic robustness analysis using a 2D vector field instance from the $\ME$ dataset (see~\autoref{sec:ME} for details). 
We study the robustness of critical points  that represent the centers of large-scale eddies. 
In (A), we visualize cancellation partners in computing the classic robustness. 
In (B), we visualize the critical points with radii proportional to their classic robustness values. 
Specifically, a number of centers (\eg, $x_0, x_1, \dots, x_{9}$) are shown to share the same lowest zero degree ancestor in the merge tree (C), thus, they are grouped together and have the same robustness value of $0.43$.
In other words, for any value $r < 0.43$, these critical points may not be canceled based on Lemma~\autoref{lemma:degree}. 
Specifically, at $r = 0.13$, the sublevel set $\Xspace_{0.13}$ contains these centers in isolation, see (C) and (D). 
Such a phenomenon happens for two reasons. 
First, some of these critical points represent centers of large-scale eddies and are surrounded by flows of a large magnitude. 
Imagining that these centers are sitting at the bottoms of deep wells (of the vector magnitude field), a large amount of perturbation is then needed to cancel these centers, and therefore they have high robustness values. 
Second, the sublevel set $\Xspace_{0.43}$ is shown to intersect significantly with the domain boundary in (E), and some of these critical points become cancellation partners due to the boundary effect. 

To mitigate these issues, we introduce the notion of  multilevel robustness. 
Roughly speaking, for a critical point $x \in \Xspace$, we define its multilevel robustness as a sequence of robustness values computed from its neighborhoods of increasing radii. 
Formally, let $B_x(a)$ denote a ball of radius $a$ surrounding a critical point $x \in \Xspace$, that is, 
$B_x(a) := \{x' \in \Xspace \mid ||x-x'|| \leq a\},$
where $||x-x'||$ represents the Euclidean distance between two points. 
The multilevel robustness of $x$ is a function 
\[R_x: [0,\infty) \to \Rspace,\]
where $R_x(a)$ is the (classic) robustness of $x$ computed {\wrt} the domain $B_x(a)$ for $a \in [0,\infty)$. 

\begin{figure}[t]
    \centering
    \includegraphics[width=\columnwidth]{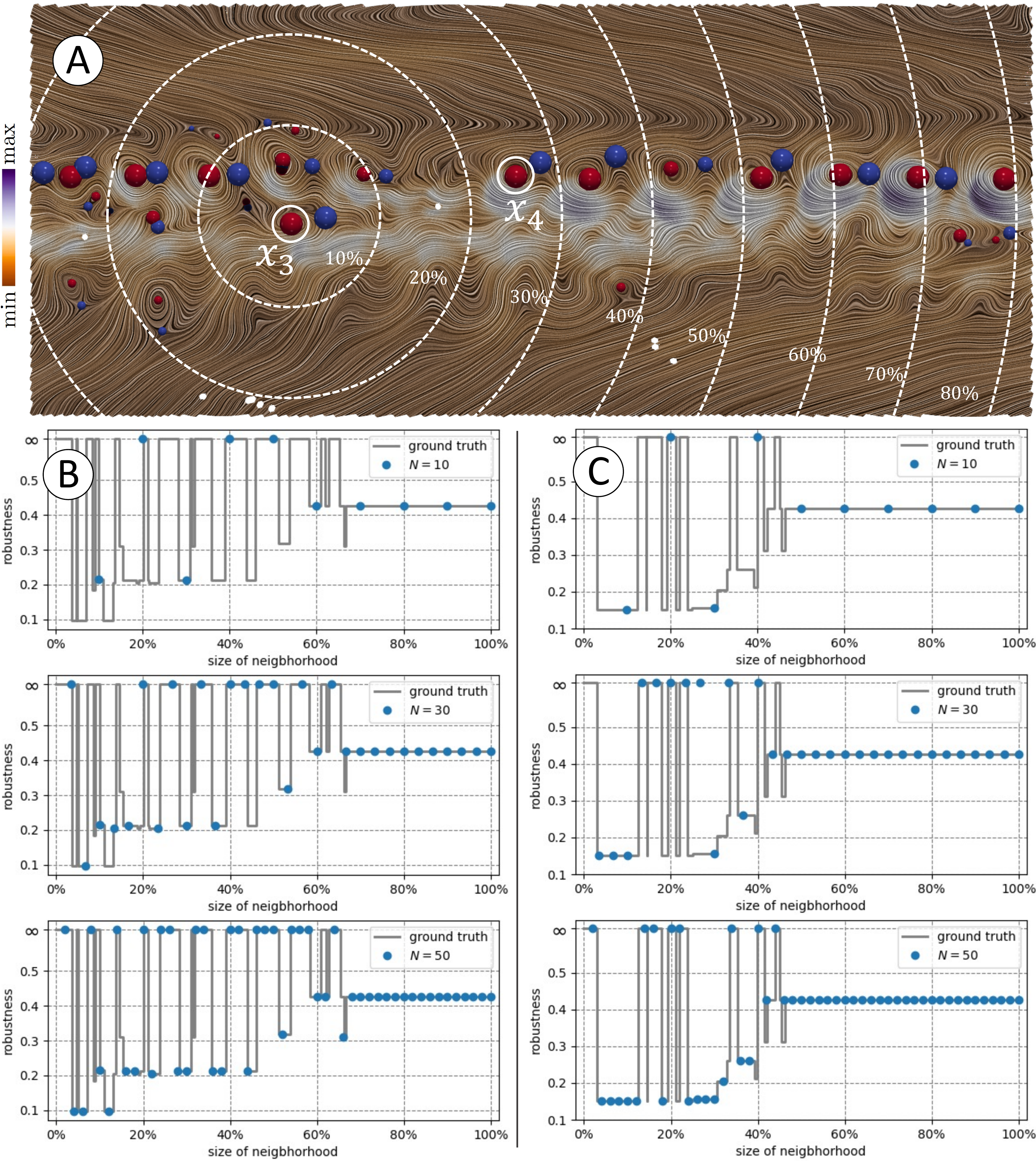}
    \vspace{-6mm}
    \caption{Multilevel robustness calculation of critical points in a 2D vector field from the $\ME$ dataset. (A) Multilevel neighborhoods for critical point $x_3$. Columns (B) and (C): $N$-level robustness and ground-truth robustness  of critical points $x_3$ and $x_4$.} 
    \label{fig:classic-robustness2}
    \vspace{-2mm}
\end{figure} 

We compute $R_x$ at a discrete number of radii. 
Assuming the domain $\Xspace$ contains $n$ critical points, then for a fixed critical point $x \in \Xspace$, as $a$ increases, its multilevel robustness will change at most $n-1$ times, since $x$ gets one more candidate of the cancellation partner as $B_x(a)$ passes through each critical point. 
Computing the multilevel robustness $R_x$ exactly (considered as the ground truth) takes $O(n^2)$ time, which is impractical for complex data with a large number of critical points.
Therefore, in practice, we approximate $R_x$ by sampling a number ($N$) of radii.
    
\autoref{fig:classic-robustness2}(A) illustrates our method in calculating the multilevel robustness. 
For a critical point $x$, we consider $N=10$ number of its neighborhoods at radius $\{a_0, \dots, a_{N-1}\}$, where each $a_i := L \times (i+1)/N$ for $L$ being the diameter of the domain $\Xspace$ (i.e.,~$L$ is the least upper bound of the set of all distances between pairs of points in the domain). 
\autoref{fig:classic-robustness2}(A) shows the $N$ neighborhoods of a critical point $x_3$ at radii at $10\%, 20\%, \dots, 100\%$ of $L$, respectively. 
At each fixed level $a_i$, we compute the classic robustness of $x_3$, giving rise to its multilevel robustness $R_{x_3}$.  

We investigate multilevel robustness for critical points $x_3$ and $x_4$ as $N$ increases, see~\autoref{fig:classic-robustness2} (B) and (C) for $N = 10, 30$ and $50$, respectively. 
Not surprisingly, $R_x(a)$ becomes a better approximation of the ground truth as $N$ increases. 
On the other hand, $R_x(a)$ appears to converge to the ground truth when $N \leq 50$ for our datasets of interests. 
Therefore, we use $N=50$ to compute multilevel robustness in the remainder of this paper.

There are a few benefits of using multilevel robustness $R_x$ for a critical point $x \in \Xspace$. 
First, $R_x$ is better at differentiating different behaviors of critical points in terms of their multiscale stability. 
As shown in \autoref{fig:classic-robustness2}, critical points $x_3$ and $x_4$ now exhibit different behaviors using $R_x$. 
Second, statistical information, such as minimum, median, and maximum of $R_x$, could be used in analysis and visualization tasks. 
Specifically, for the remainder of this paper, we work with the minimum of $R_x$ for critical point tracking, selection, and comparison, which is defined as 
\[
\minR_{x} := \min_{a \in [0,~L)} R_x(a). 
\] 
$\minR_x$ captures the smallest possible robustness of $x$ with varying neighborhood sizes, and thus alleviates the artifacts induced by the boundary effects in classic robustness calculation.  
In addition, $\minR_x$ is shown to be highly correlated with physical quantities employed by domain scientists who study tropical cyclones; compare~\autoref{fig:HK} for a concrete example. 

%% file: sec-visualization.tex
\section{Method: Multilevel Robustness for Visualization Tasks}
\label{sec:vis}

With the newly introduced multilevel robustness framework, we develop its usage in visualization tasks. 
Such a new notion of robustness can be combined seamlessly with any feature tracking algorithm. 
We choose to integrate it with  FTK~\cite{GuoLenzXu2021}, a state-of-the-art feature tracking technique. 
In particular, we demonstrate that the minimum multilevel robustness $\minR_{x}$ can be integrated with the FTK algorithm to improve \myupdate{the original FTK} feature tracking and selection results for scientific simulations.  

\begin{figure}[t]
    \centering
    \includegraphics[width=0.98\columnwidth]{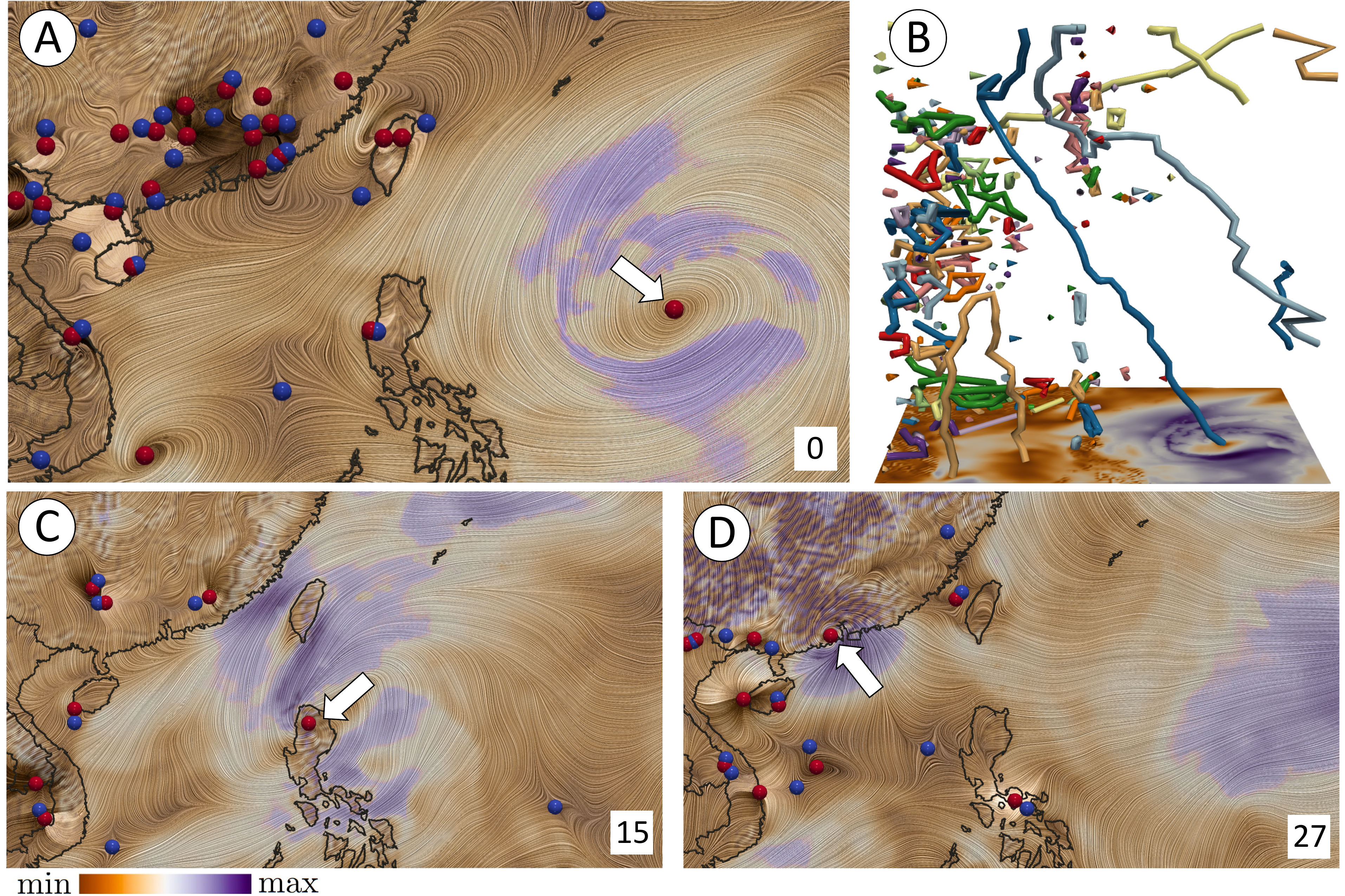}
    \vspace{-2mm}
    \caption{\myupdate{Critical points and FTK tracking results for the $\EW$ dataset. (A), (C), and (D): selected vector fields with their critical points. (B) Feature (i.e., critical point) tracking results using FTK; each trajectory is colored by the trajectory id.}} 
    \label{fig:VisExample}
    \vspace{-2mm}
\end{figure}

\para{Illustrative dataset.} In this section, we use an $\EW$ dataset to illustrate our method. 
$\EW$ is a time-varying 2D vector field processed using a HiResMIP-v1.0 (1950-Control) dataset~\cite{CaldwellMametjanovTang2019} from the Energy Exascale Earth System Model (E3SM)~\cite{GolazCaldwellVan2019} project. 
\myupdate{It has an approximate horizontal resolution of $0.25^\circ$ (degree) in the atmosphere ($28$ km grid spacing), with an ocean and sea ice grid of $18$ km in the mid-latitudes and $6$ km at the equator and poles.}
We truncate a rectangular region around south Asia ($10$ N$^\circ$ to $30$ N$^\circ$ and $105$ E$^\circ$ to $140$ E$^\circ$), and select $36$ time steps from September 18, the 26th run of the 1950-control dataset with 6 hours as the time gap. 
We use \emph{UBOT} and \emph{VBOT} as 2D vector fields, which correspond to lowest model level zonal and meridional wind, respectively. 
These instances describe the movement of a main  cyclone, which forms in the Pacific ocean, passes through the Philippines (around time steps 15-18), makes landfall (time step 27), and dissipates (around time step 31) at the mainland of south Asia; \myupdate{ see~\autoref{fig:VisExample} (A), (C), and (D), which visualizes the vector fields associated with time steps $0$, $15$, and $27$. }
The cyclone of interest is indicated by the white arrows.

\para{Initial computation of trajectories and multilevel robustness.} 
The initial (critical point) trajectories of time-varying vector field data are computed by FTK~\cite{GuoLenzXu2021} \myupdate{(see~\autoref{fig:VisExample} (B))},  
and we then use the method of Tricoche~\etal~\cite{TricocheScheuermannHagen2001b} to calculate degrees of critical points in individual time steps. 
The computation of multilevel robustness is parallelized with Eden~\cite{SimmermanOsborneHuang2012}, which schedules and manages a number of small tasks on a high-performance computing cluster.  In our implementation, each task is associated with one critical point and a neighborhood size. As a result, the robustness computation of $n$ critical points with $N$ levels leads to $n\times N$ independent tasks. 

\subsection{Enhancing Feature Tracking with Multilevel Robustness}

\begin{figure}[t]
    \centering
    \includegraphics[width=\columnwidth]{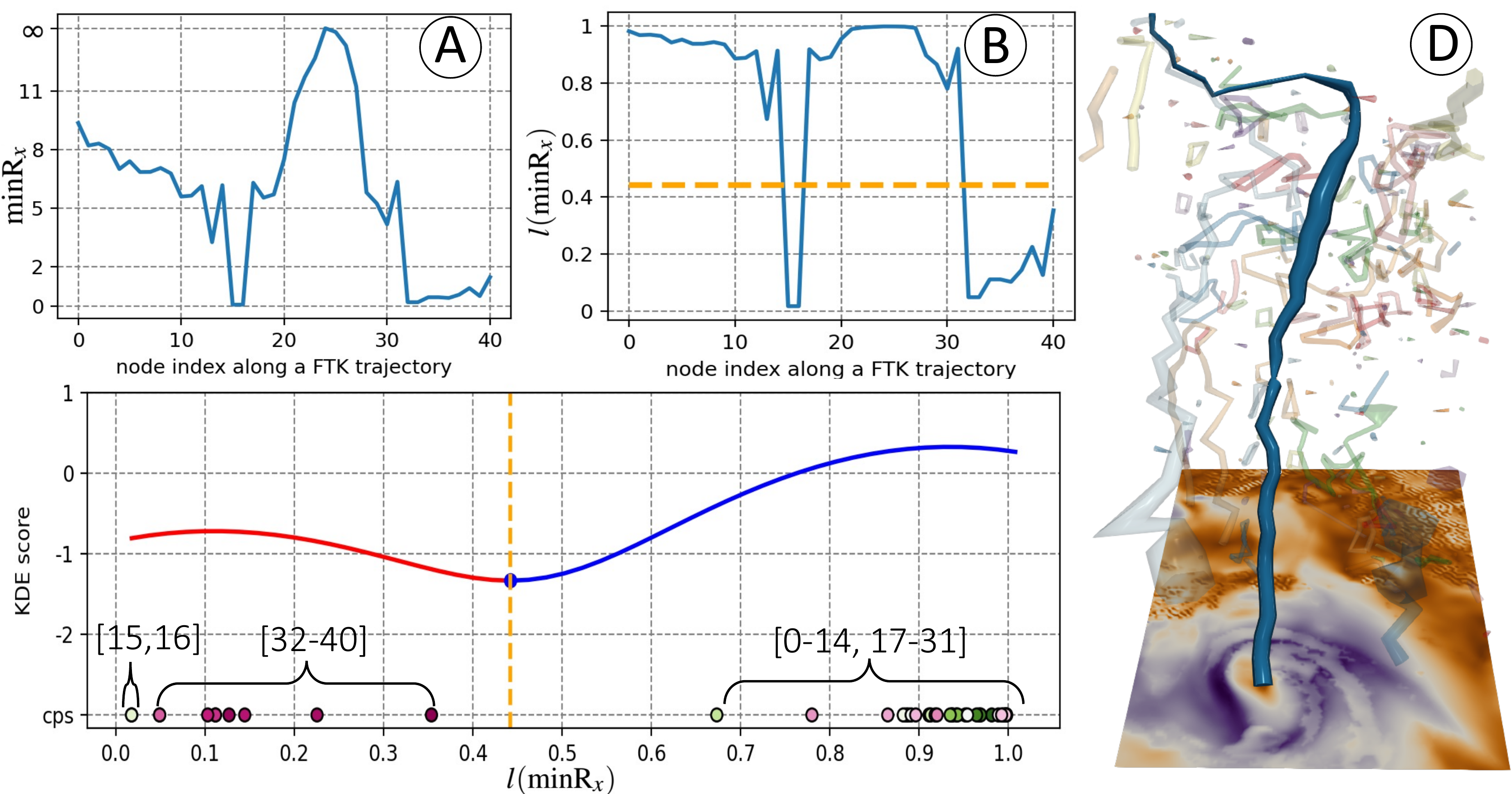}
    \vspace{-6mm}
    \caption{(A) Minimum multilevel robustness values $\minR_x$ for a critical point $x$ along the selected FTK trajectory. (B) Logistic transformation of $\minR_x$ along the FTK trajectory. (C) Segmentation of points in the trajectory based on kernel density estimation. (D) The selected FTK trajectory is highlighted in solid blue, whereas other trajectories are transparent.} 
    \label{fig:VisExample-Clustering}
    \vspace{-2mm}
\end{figure}

In this section, we show that multilevel robustness -- in particular, the minimum multilevel robustness $\minR_x$ -- can significantly improve the feature tracking results. 
Given the initial trajectory of a critical point $x$ together with its multilevel robustness $R_x$ over time, we may visualize the trajectory by encoding the statistical information of $R_x$ along the trajectory, such as its minimum, median, and maximum values.
\autoref{fig:VisExample-Clustering}(D) shows a visualization of these trajectories where the radius of each point along a trajectory is shown to be  proportional to the minimum of its multilevel robustness $\minR_x$. 

The main idea of feature tracking with multilevel robustness is to segment the initial trajectories (obtained by FTK or any other feature tracking algorithms) into multiple pieces with similar robustness values. 
As discussed in~\autoref{sec:ml-robustness}, we prefer to use the minimum of multilevel robustness $\minR_x$ to quantify the stability of critical points, which alleviates the artifacts introduced by the boundary effect in classic robustness computation. 
We demonstrate that our tracking strategy improves the initial FTK trajectories and captures stable features in the domain, for example, in tracking the main cyclone for the $\EW$ dataset. 

We focus our analysis on an FTK trajectory that contains the main cyclone. The blue trajectory in~\autoref{fig:VisExample-Clustering}(A and D) shows the $\minR_x$ values along the trajectory.  
Note that in \autoref{fig:VisExample-Clustering}(A),  each trajectory is a parameterized curve, where an integer index (horizontal axis) corresponds to the parameter used in the parameterization; thus, each trajectory is not necessarily monotonic in time. 
\myupdate{In the remaining of this section, we use indices to refer to nodes along a trajectory.}
As shown in~\autoref{fig:VisExample-Clustering}(A), $\minR_x$ decreases significantly at \myupdate{indices} $15$ and $16$, and its value remains low after \myupdate{index} $32$. 

Our first step is to segment a given trajectory into groups of critical points with similar robustness values. 
This step is supported by the theoretical work in~\cite{SkrabaWang2014b}, where correspondences between critical points may be inferred based on their closeness in robustness. 
 To induce a segmentation more easily, we can amplify the signal $\minR_x$ with a logistic transformation.
 Starting from a standard logistic function $s(z) = {1}/\left({1+e^{-k (z - z_0)}}\right)$, set $z = \minR_x$ at a fixed time step and $z_0 = 0$ (the minimal possible robustness value).  Since $\minR_x \in [0, \infty)$, we have $s(z) \in [1/2, 1]$. 
 Introducing a normalization term, we have  
 \[
 l(\minR_x)=\frac{2}{1+e^{-k\cdot \minR_x}}-1,
 \] 
so that $l(\minR_x) \in [0,1]$. 
Here, $k$ is the logistic growth rate or the steepness of the curve of the function. 
We set $k=0.5$ for most cases, and discuss the parameter choices later. 
There are two justifications for using a logistic transformation. 
First,  $\minR_x$ may be infinity when a critical point $x$ cannot find a cancellation partner in the known data domain; $l(\minR_x)$ is thus constrained within the range $[0,1]$, making parameter selection easier. 
Second, $l(\minR_x)$ is less sensitive {\wrt} to the changes in $\minR_x$, and therefore it focuses only on significant changes of $\minR_x$.  
For example, as shown in \autoref{fig:VisExample-Clustering}(A) and (B), the logistic transformation $l$ maps the $\minR_x \in [0, \infty]$ to $l(\minR_x) \in [0,1]$. 
Furthermore, it helps to differentiate unstable critical points along the trajectory from relatively stable ones. 
As shown in~\autoref{fig:VisExample-Clustering}(B), there appears to be clear separations between indices $[15, 16]$ and $[31, 40]$ from the rest of the trajectory. 

Our second step is to cluster critical points along a trajectory into different groups using $l(\minR_x)$ and kernel density estimation (KDE) with a Gaussian kernel. 
As illustrated in~\autoref{fig:VisExample-Clustering}(C), by choosing an appropriate bandwidth parameter $\sigma$ for the Gaussian kernel, we can further cluster the critical points with indices $[15,16]$ and $[32, 40]$ from those with indices $[0,14]$ and $[17, 31]$.   
$\sigma$ controls the smoothness of a KDE, where a small $\sigma$ leads to more segments. 
For our experiments, we set $\sigma=0.2$ as the default value; see a later section for parameter tuning. 

\begin{figure}[!t]
    \centering
    \includegraphics[width=\columnwidth]{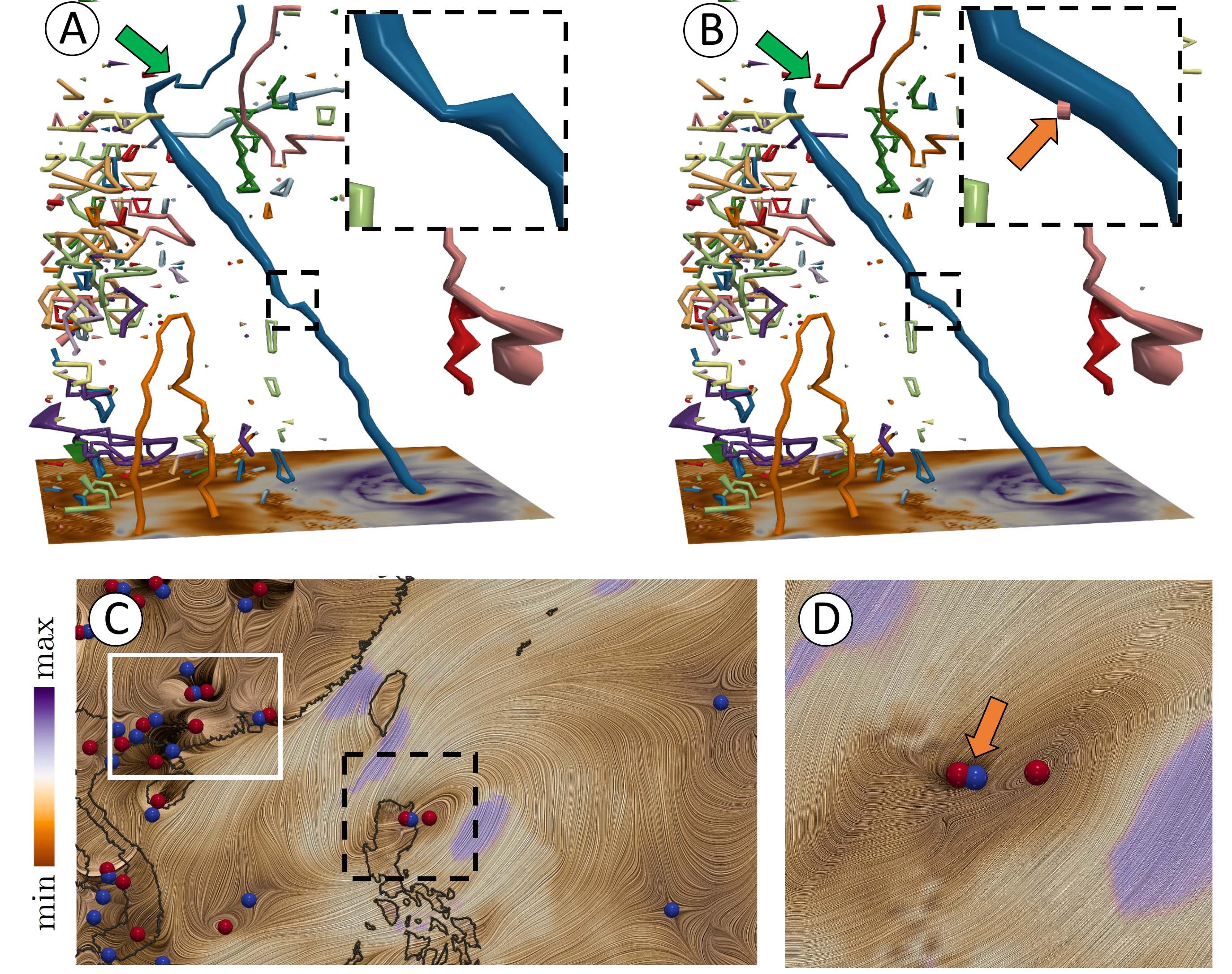}
    \vspace{-6mm}
    \caption{Comparing feature tracking results for a time-varying $\EW$ dataset. (A) Initial (classic) FTK trajectories; (B) trajectories obtained based on multilevel robustness, with zoomed-in view for indices  13-15. The vector field at index 14 is visualized in (C) with a zoomed-in view around the Philippians in (D).} 
    \label{fig:VisExample-Tracking}
    \vspace{-2mm}
\end{figure}

\begin{figure*}[t]
    \centering
    \includegraphics[width=2.01\columnwidth]{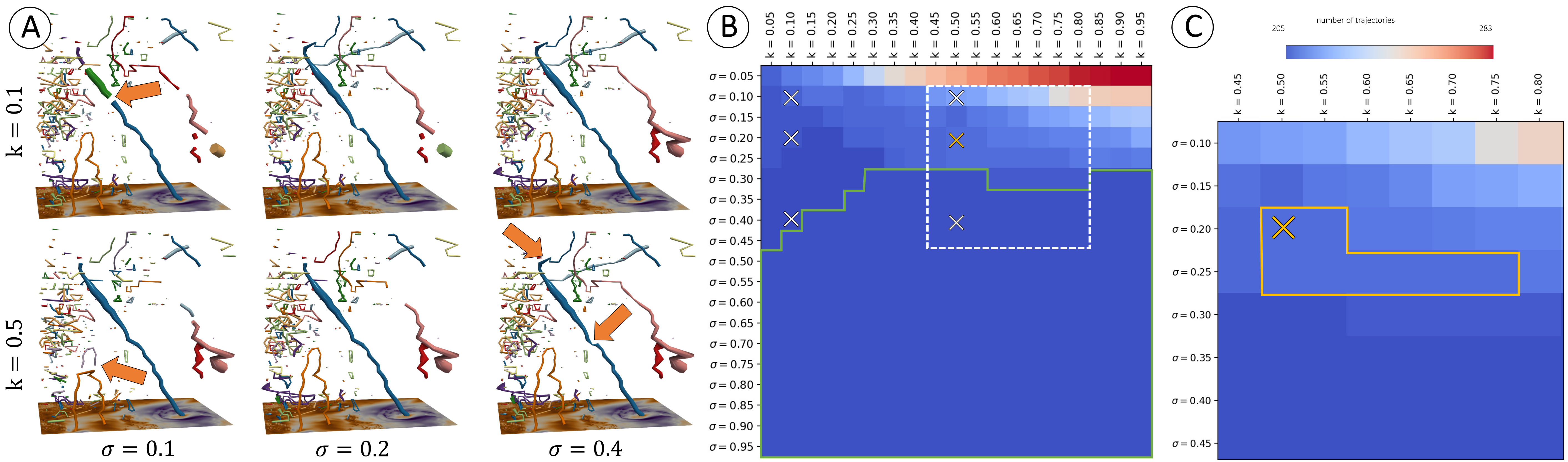}
    \vspace{-1mm}
    \caption{Parameter selection of $k$ and $\sigma$ for the $\EW$ dataset. (A) From top to bottom, feature tracking results for $k=0.1$ and $0.5$. From left to right, $\sigma=0.1$, $0.2$, and $0.4$. \myupdate{(B) Number of trajectories {\wrt}~parameter selection of $k$ and $\sigma$. (C) A zoomed-in view of (B) for the region surrounded by the dashed white box.}} 
    \label{fig:VisExample-parameter}
    \vspace{-2mm}
\end{figure*}
Our third step is to refine critical point trajectories based on the clustering results. 
Critical points belonging to the same cluster are reconstructed as a new trajectory by examining spatial faces and spacetime edges~\cite{GuoLenzXu2021}.
As illustrated in~\autoref{fig:VisExample-Tracking}(B), the selected blue FTK trajectory is segmented into three pieces: a orange trajectory connecting critical points of indices $[15,16]$, an red trajectory connecting critical points with indices $[31, 40]$, both with small robustness values; and the remaining blue trajectory with large robustness values.  
In particular, the new blue trajectory in (B) is reconstructed by connecting critical points of indices $14$ and $16$ following the approach in~\cite{GuoLenzXu2021}.

Based on domain knowledge, a critical point representing the center of a cyclone should have a high stability measure across time before it hits the land and dissipates. 
Take a close look at the trajectory at index 14. As shown in \autoref{fig:VisExample-Tracking} (D), there are two critical points with low stability near the landmass of the Philippines. 
Since the classic FTK algorithm only considers the correspondences of critical points based on 0-levelset extraction, these two critical points are included in the initial trajectory in ~\autoref{fig:VisExample-Tracking}(A). 
Furthermore, the critical points with indices $[31, 40]$ are likely unstable features when the cyclone makes landfall and dissipates.

Our feature tracking method is used as a postprocessing step to segment initial FTK trajectories into more meaningful segments, based on multilevel robustness. 
In particular, we compare the initial trajectory with our new trajectory based on multilevel robustness in~\autoref{fig:VisExample-Tracking}(A)-(B). 
The initial trajectory includes a pair of critical points on the island of the Philippines with low robustness values; whereas our method successfully tracks the main cyclone and removes these two critical points from the main trajectory, as indicated by an orange arrow in \autoref{fig:VisExample-Tracking}(B). 
Also, our method splits the initial trajectory into a red trajectory when the cyclone hits the south of Asia, as indicated by the green arrow in \autoref{fig:VisExample-Tracking}(B), since the low robust tail of the initial trajectory is formed by unstable features on land; see critical points within the white box of~\autoref{fig:VisExample-Tracking}(C).

\para{Parameter selection for $k$ and $\sigma$.} 
We now discuss how the choice of $k$ from the logistic transformation and bandwidth $\sigma$ from the KDE affect the feature tracking results. 
$k$ is used to control the growth rate of the logistic transformation. 
\autoref{fig:VisExample-parameter}\myupdate{(A)} shows the feature tracking results for $k=0.1$ and $0.5$. 
When $k$ is relatively small (\eg, $k=0.1$), the logistic transformation cannot differentiate stable features from unstable ones, regardless of the values of $\sigma$. 
For example, for $k=0.1$ and $\sigma=0.1$, our method over-segments the initial trajectory. 
As $\sigma$ increases (\autoref{fig:VisExample-parameter}\myupdate{(A)} 1st row), our method does not exclude unstable features on the island and those on the mainland.  
On the other hand, we obtain reasonable (similar) feature tracking results for $k \in [0.3, 1.0]$. 
This indicates that a slightly higher value of $k$ is effective in differentiating stable and unstable features. Therefore, we set $k=0.5$ for our experiments in~\autoref{sec:results}.

For a fixed $k$ value, \autoref{fig:VisExample-parameter}\myupdate{(A)} also shows the feature tracking results for $\sigma=0.1$, $0.2$, and $0.4$, respectively.  
As discussed previously, a small $\sigma$ will likely introduce the over-segmentation of a given trajectory. 
For example, when $\sigma=0.1$, the trajectory representing a merging behavior of a pair of critical points on the left bottom corner of $\EW$ is divided into two parts, as indicated by the orange arrows in \autoref{fig:VisExample-parameter}\myupdate{(A)} (1st column and 2nd row). 
When $\sigma$ is large, the KDE curve becomes too smooth to differentiate stable and unstable features. As shown in \autoref{fig:VisExample-parameter}\myupdate{(A)} (3rd column and 2nd row), if we set $\sigma=0.4$, the blue trajectory is similar to the trajectory using classic FTK algorithm. 
This means that our feature tracking under $\sigma=0.4$ fails to extract the main cyclone from other unstable features around the regions indicated by orange arrows. 
Finally, since both KDE and $l(\minR_x)$ have a range of $[0,1]$, we set $\sigma=0.2$ as default since it works well in most cases considered in~\autoref{sec:results}.

\myupdate{Additionally, we use a heatmap that records the number of trajectories under various parameter settings in \autoref{fig:VisExample-parameter}(B), which provides a supplementary view for parameter selection. 
The values of $k$ and $\sigma$ from \autoref{fig:VisExample-parameter}(A) are highlighted by crosses in \autoref{fig:VisExample-parameter}(B).
For parameter selection, we look for the regions where the number of trajectories remains relatively stable with a range of values for $k$ and $\sigma$.    
For example, when $\sigma \geq 0.45$, our   framework produces the same number of trajectories as the original FTK tracking result. It means that with a relatively high $\sigma$, our framework cannot differentiate between stable and unstable features, e.g., see the region surrounded by the green boundary in \autoref{fig:VisExample-parameter}(B). 
On the other hand, a small $\sigma$ tends to over-segment the initial FTK tracking results, see the first row of \autoref{fig:VisExample-parameter}(B).
Our default values of $k$ and $\sigma$ come from the region surrounded by orange boundary in \autoref{fig:VisExample-parameter} (C). Any combination of $k$ and $\sigma$ from this region leads to the same post-processed feature tracking result.}

\begin{figure}[!t]
    \centering
    \includegraphics[width=\columnwidth]{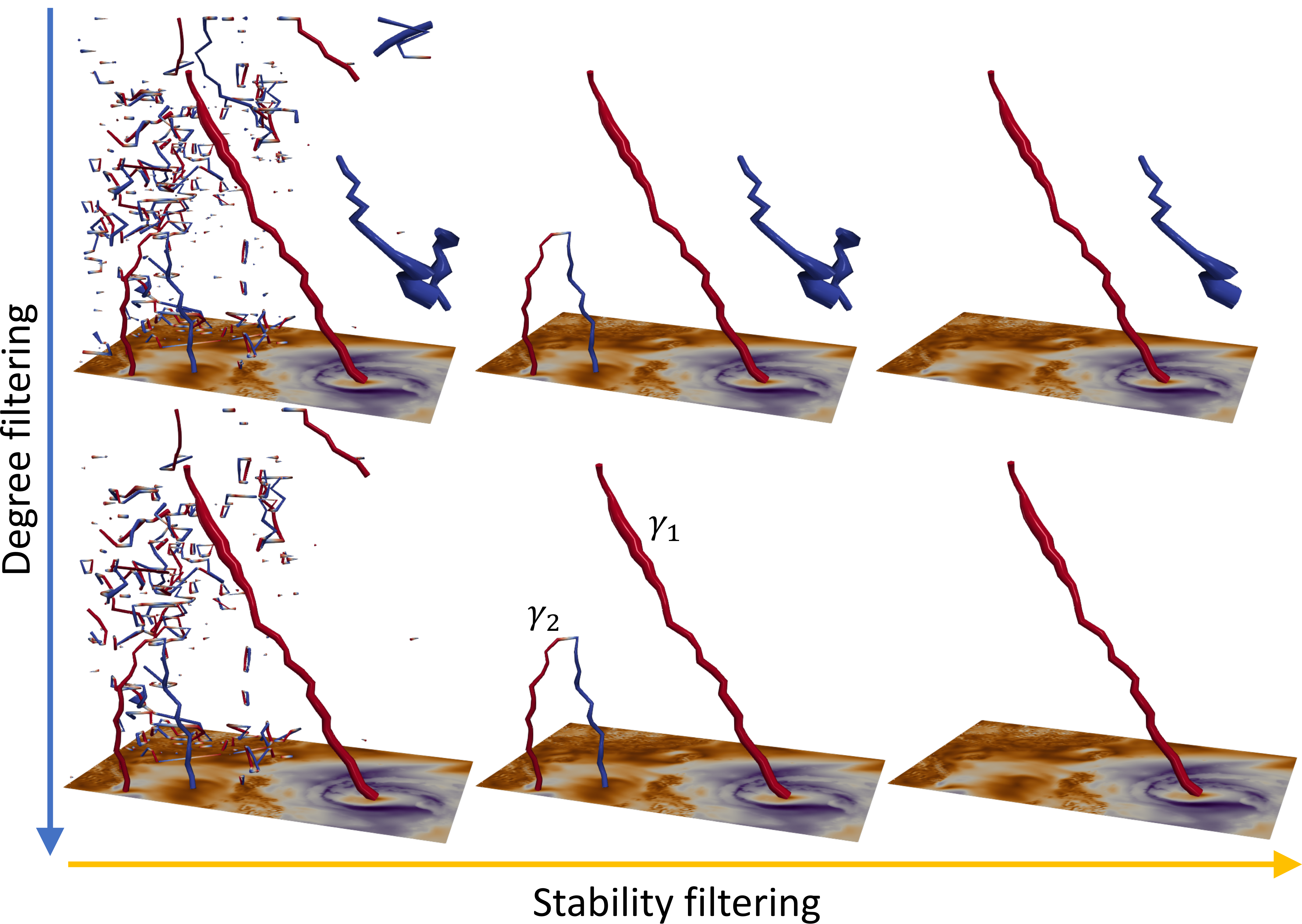}
    \vspace{-4mm}
    \caption{Feature selection for the $\EW$ dataset. 
    From top to bottom: degree filtering with thresholds at $-1$ (\ie, keeping all saddles) and $-0.2$ respectively. 
    From left to right: stability filtering with thresholds at $0$, $0.2$, and $0.4$, respectively.  
     Trajectories are colored by the degrees of critical points, where red means $+1$ and blue means $-1$. The radius of a trajectory is proportional to $\minR_x$.}
    \label{fig:VisExample-Selection}
    \vspace{-2mm}
\end{figure}

%

\subsection{Feature Selection with Multilevel Robustness}

This section demonstrates feature selection aided by multilevel robustness. We introduce two filters, one based on $\minR_x$, and the other based on degree information. 
Both filters help to reduce visual clutter and highlight dominant features in the domain. 

Our first feature selection strategy is referred to as the \emph{stability filtering}. 
For any trajectory, this strategy considers the topological notion of stability in terms of $l(\minR_x)$, as well as its temporal stability in terms of the lifespan. 
Let $\gamma$ denote a parameterized trajectory,  $|\gamma|$ is its total length. 
Formally, we define the \emph{stability measure} of a trajectory $\gamma$ as follows:
\begin{equation}
\label{eqn:stability}
  b(\gamma) := \frac{\sum_{x \in \gamma}l(\minR_x)}{|\gamma|} \cdot \frac{t_\gamma}{T},
\end{equation}
where $T$ is the temporal span of all trajectories (\eg, $T=36$ for the $\EW$ dataset), and $t_\gamma$ is the temporal span of $\gamma$ in terms of the maximum difference between node indices. 
The first term in Eqn.~\eqref{eqn:stability} captures the average pointwise stability (in a logistic scale), whereas the second term encodes the lifespan of the trajectory. 
By definition, $b(\gamma)$ has a range in $[0, 1]$. 

Our second feature selection strategy is referred to as the \emph{degree filtering}. 
That is, we select trajectories based on their pointwise average degree. 
Formally, for a trajectory $\gamma$, its \emph{average degree} is 
\begin{equation}
d(\gamma) := \frac{\sum_{x \in \gamma} \mydeg(x)}{|\gamma|}, 
\end{equation}
where $\mydeg(x)$ is the degree of a critical point $x \in \gamma$. 
Since a critical point may be of degree $+1$ or $-1$, $d(\gamma)$ has a range in $[-1, 1]$.  
For our experiments involving cyclones and ocean eddies, we work primarily with critical points with a degree of $+1$, which correspond to centers of cyclones and eddies.  
Domain scientists mainly care about centers in our applications.
For example, the trajectory that represents the main cyclone in the $\EW$ dataset contains critical points (centers) whose degrees are all $+1$. 
Therefore, trajectory $\gamma_1$ within \autoref{fig:VisExample-Selection} has $d(\gamma_1) = +1$. For trajectory $\gamma_2$, $d(\gamma_2) = 0$, since critical points on its left branch have degrees $+1$ and critical points on its right branch have degrees $-1$.  
Once a trajectory is enriched with a stability measure and an average degree, one may select features based on these criteria jointly or independently. 
As illustrated in \autoref{fig:VisExample-Selection}, we successfully selected the trajectory that represents the main cyclone with degree filtering and stability filtering.

%% file: sec-results.tex
\section{Results with Large-Scale  Simulations}
\label{sec:results}
\begin{figure*}[!t]
    \centering
    \includegraphics[width=2\columnwidth]{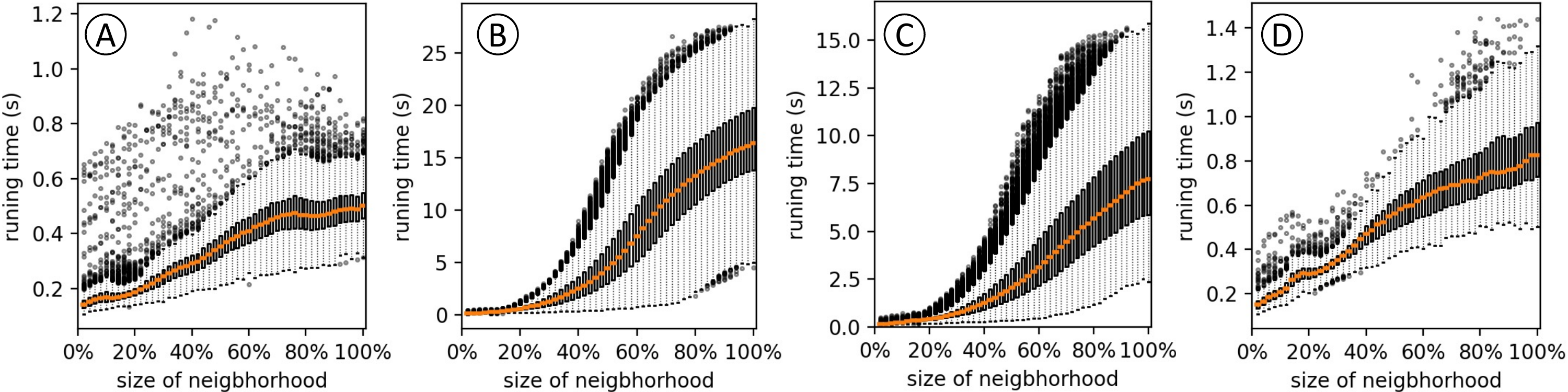}
    \caption{Boxplots of running time with different neighborhood sizes for the (A) $\EW$, (B) $\EWL$, (C) $\HK$, and (D) $\ME$ datasets. } 
    \label{fig:runningTime}
    \vspace{-2mm}
\end{figure*}

\begin{figure*}[!t]
    \centering
    \includegraphics[width=1.0\linewidth]{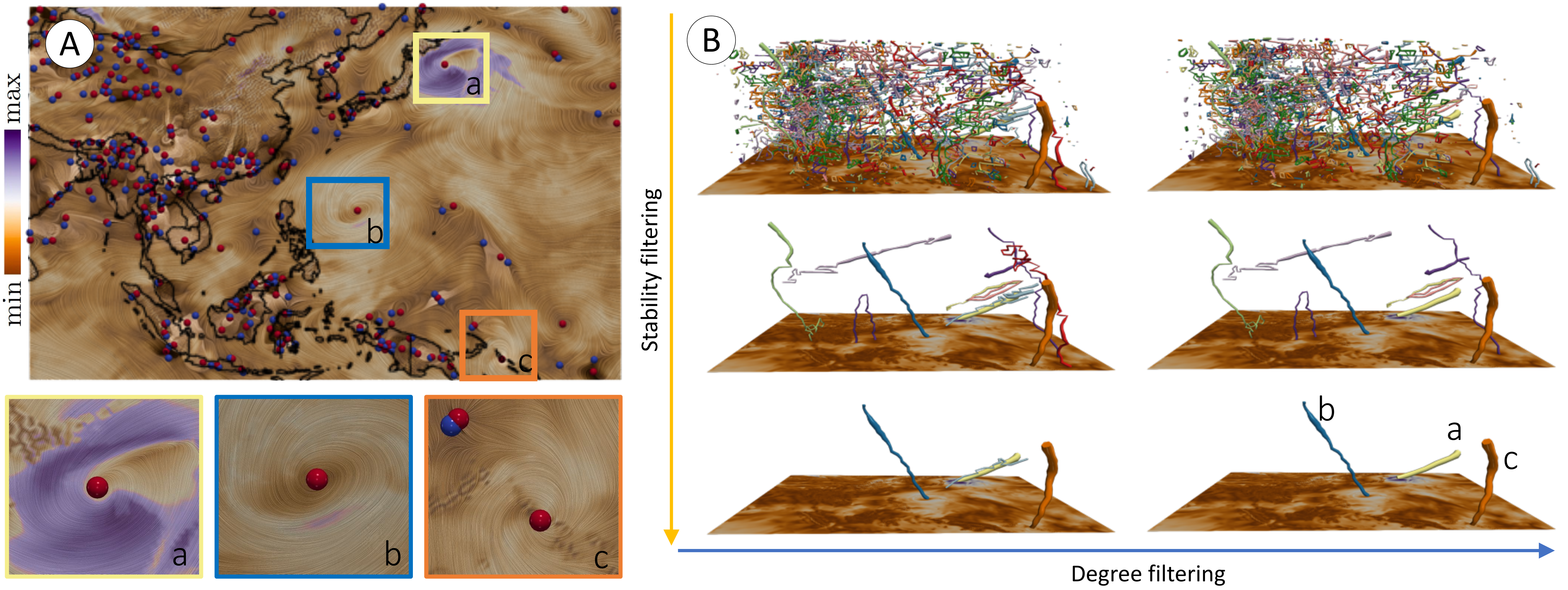}
    \vspace{-4mm}
    \caption{Feature tracking and selection for the $\EWL$ dataset. (A) A selected vector field with zoomed-in views, where
saddles are in blue, and critical points with degree $+1$ are in red.
(B) Tracking results with degree filter thresholds at $-1$ and $-0.2$ (from left to right) and stability filter thresholds at $0$, $0.2$, and $0.4$ (from top to bottom).} 
    \label{fig:EWL}
    \vspace{-2mm}
\end{figure*}

\begin{figure*}[!t]
    \centering
    \includegraphics[width=1.0\linewidth]{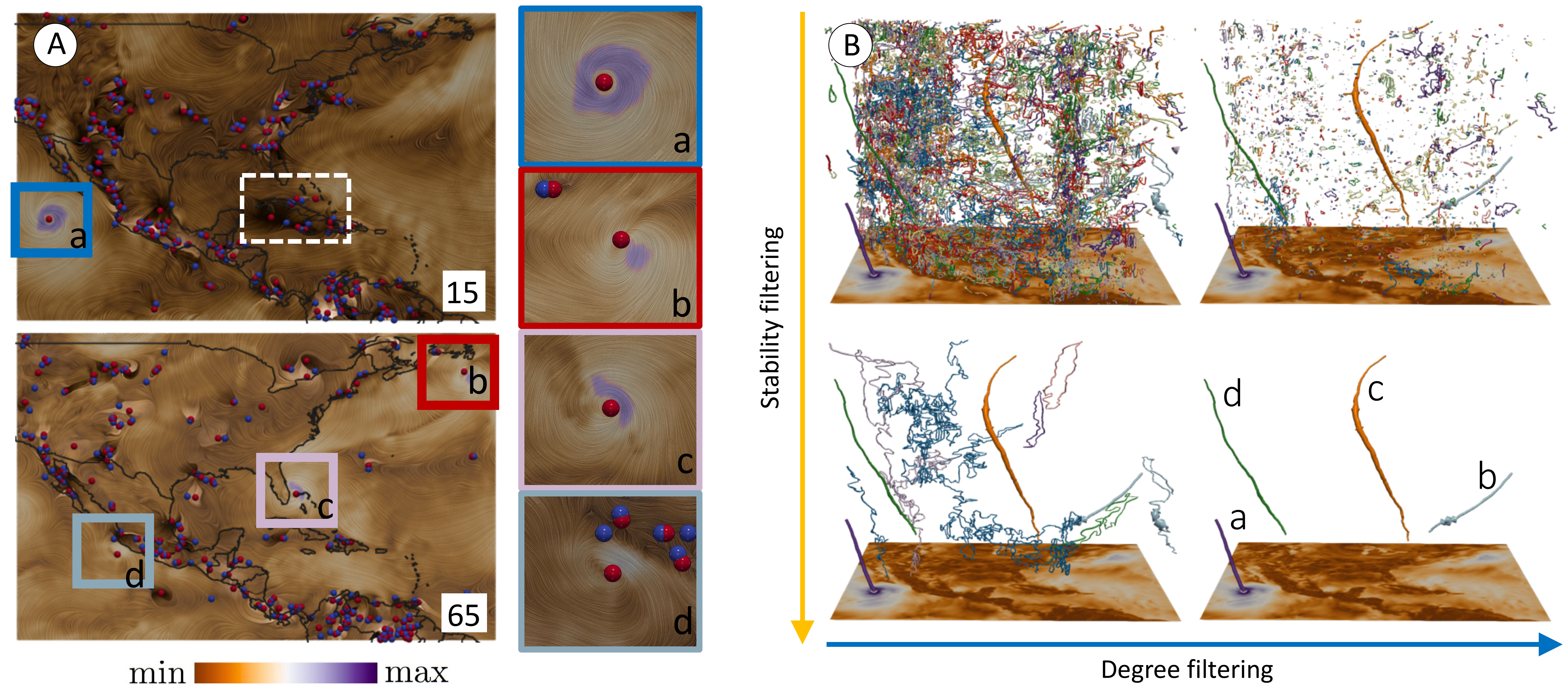}
    \vspace{-4mm}
    \caption{Feature tracking and selection for the $\HK$ dataset. (A) Selected vector fields with zoomed-in views.
(B) Tracking results with stability filter thresholds at  $0$ and $0.1$ (from left to right) and degree filter thresholds at $-1$ and $-0.02$ (from top to bottom).}
    \label{fig:HK}
    \vspace{-2mm}
\end{figure*}

\begin{figure}[!t]
    \centering
    \includegraphics[width=\columnwidth]{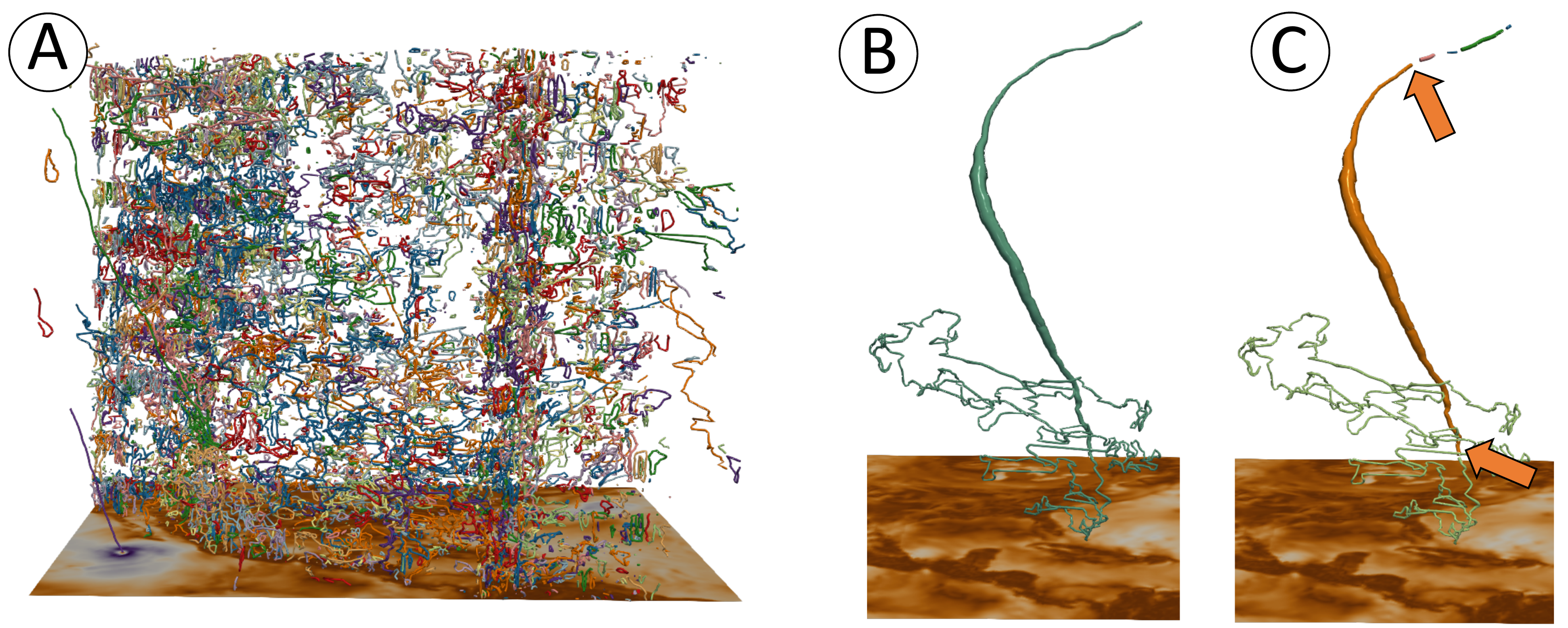}
    \vspace{-6mm}
    \caption{Feature tracking result for the $\HK$ dataset. (A) Initial FTK trajectories. (B) Initial FTK trajectory that contains \emph{Katrina}. (C) Segmenting trajectory in (B) with multilevel robustness; new segments are assigned different colors.} 
    \label{fig:HK-mainTrajectory}
    \vspace{-2mm}
\end{figure}

We demonstrate the use of multilevel robustness in feature tracking, selection, and comparison for large-scale scientific simulations.  
\autoref{tab:data} lists some basic information for datasets used in this paper, including the number of time steps, grid nodes, grid cells (triangles), and critical points. 
We also provide a brief running time analysis for all datasets based on each task discussed in \autoref{sec:vis} (i.e.,~a single task involves computing classic robustness of a given critical point at a fixed radius). 
All tasks are arranged on a cluster with 664 nodes (128GB DDR4 and 36 cores). We utilize 16 nodes in all experiments, which means that at most $16 \times 36=575$ tasks can run at the same time.  
Since the running time of each task is highly correlated with the size of neighborhood in the robustness calculation, \autoref{fig:runningTime} provides the box plots of running time at each level of robustness for all critical points in a given dataset; also see \autoref{tab:data} (last column) for the range of running time of tasks for each dataset.

In the following, all timestamps in the descriptions are represented in the universal coordinated time (UTC).

\begin{table}[t]
  \caption{Datasets used in this paper.  \#steps, \#nodes, \#cells, and \#CPs, respectively, mean the number of timesteps, number of vertices in the mesh, number of cells in the mesh, and number of critical points.  Running times are in seconds.} 
  \label{tab:data}
  \scriptsize%
	\centering%
  \begin{tabular}{cccccc
	}
\hline
     \textbf{Dataset }& \textbf{\#steps} &   \textbf{\#nodes}  &   \textbf{\#cells} &   \textbf{\#CPs} & \pbox{15cm}{\textbf{time per task}}\\
     \hline
     $\EW$&36&13,639&26,356&8-53&0.11-1.18\\
     $\EWL$&36&99,007&195,602&238-425&0.11-27.72\\
     $\HK$&216&50,861&100,800&208-501&0.11-15.97\\
     $\ME$&$20\times 4$&37,929&75,063&18-65&0.11-1.44\\
\hline
  \end{tabular}%
\end{table}

\subsection{Feature Tracking and Selection}

\para{E3SM Wind L dataset.} 
We revisit the HighResMIP-v1.0 (1950-Control) dataset~\cite{CaldwellMametjanovTang2019} from 
E3SM simulations. 
Instead of truncating a small region for illustrative purposes in~\autoref{sec:vis}, we enlarge the $\EW$ dataset to the $\EWL$ dataset by choosing the region from $10$ S$^\circ$ to $50$ N$^\circ$ for latitude and from $80$ E$^\circ$ to $175$ E$^\circ$ for longitude. 

As shown in~\autoref{fig:EWL}(B), with the appropriate trajectory segmentation, stability filtering, and degree filtering, our framework detects the trajectories of three main cyclones in the domain, denoted as $a$, $b$, and $c$.  
Trajectory $a$ appears at the east of Japan from time step $0$, moves to the east, and disappears on the right boundary at time step $9$. 
Trajectory $b$ exists from time step $0$ to $31$, and coincides with the selected cyclone trajectory of $\EW$ in~\autoref{fig:VisExample-Tracking} and~\autoref{fig:VisExample-Selection}.  	
Trajectory $c$ stays on the right bottom corner of the domain from time steps 0 to 21. 
For further investigation of these trajectories, we visualize the vector fields associated with time step 0 in~\autoref{fig:EWL}(A), with color map based on the magnitude of the vector fields.  We also give the zoomed-in views of the detected main features.

\para{Hurricane Katrina dataset.} 
\emph{Hurricane Katrina} was a large and destructive Category 5 Atlantic hurricane that formed on August 23, 2005, and dissipated on August 31, 2005. 
Our $\HK$ dataset is truncated from ECMWF Reanalysis v5 (ERA5), which is produced by the Copernicus Climate Change Service (C3S)~\cite{C3S}.
ERA5 provides hourly estimates of the global climate information covering the period from January 1950 to the present with the spatial grid resolution of 30 km. The rectangular region is centered at the southeast of the contiguous U.S. ($5$ N$^\circ$ to $50$ N$^\circ$ and $120$ W$^\circ$ to $50$ W$^\circ$); the time steps range from 12:00, August 23, 2005, to 23:59, August 31, 2005. 
Since ERA5 uses a one-hour time gap, our $\HK$ dataset contains $9 \times 24 = 216$ instances. 
We choose 10m zonal and meridional wind speed as the 2D vector field, since in the near-surface the hurricane core represents a region of strong convergence and associated vertical motion.

We start with the initial trajectories provided by FTK, shown in \autoref{fig:HK-mainTrajectory}(A).   
Due to visual clutter among thousands of trajectories, it is hard to identify the principal features. 
However, our multilevel robustness framework is able to detect four dominant features in the domain, after trajectory segmentation and filtering, shown as trajectories $a$ to $d$ in~\autoref{fig:HK}(B).  
We visualize vector fields associated with time steps $15$, $65$, and $120$, and highlight these four features in the zoomed-in views; see~\autoref{fig:HK}(A). 
In particular, trajectory $c$ contains critical points representing the center of \emph{Katrina}. 

We perform a detailed analysis of the robustness-based  segmentation of trajectory $c$. 
As shown in~\autoref{fig:HK-mainTrajectory}(B), the initial FTK trajectory containing $c$ also contains a number of critical points that are not associated with \emph{Katrina}. 
These critical points are part of the same initial trajectory with $c$. 
Multilevel robustness is used to segment this initial trajectory by differentiating spurious features from the features of interest, using $k=0.5$ for the logistic transformation and $\sigma=0.2$ for the KDE. 
As shown in~\autoref{fig:HK-mainTrajectory}(C), our approach extracts the trajectory $c$ that represents \emph{Katrina} with two segmentation points highlighted by orange arrows. 

Trajectory $c$, which represents \emph{Katrina}, exists between time steps 33 and 201, which correspond to 9:00, August 24, 2005 and 9:00 August 31, 2005. 
This means our framework does not capture \emph{Katrina} on August 23, 2005, when it was a tropical depression. 
Taking a closer look at the critical points within the dashed white box from~\autoref{fig:HK}(A) at time step 15, it is hard for us to extract the low robust critical point that can represent \emph{Katrina} as there are many unstable features nearby. 
However, this could be an artifact of the reanalysis product being used: since data assimilation in ERA5 occurs at 09:00, it is likely that the reanalysis product has been artificially adjusted to include \emph{Katrina}'s precursor. 
Overall, our framework works well in detecting \emph{Katrina} when it strengthened into a tropical storm on the morning of August 24.

Based on input from domain scientist, our hypothesis of data preparation being an issue for Katrina is related to the disparate character of the storm at the hour of its first detection and the previous hour. The storm first appears when the data assimilation system is employed to generate new initial conditions for the forecast, suggesting that its development was not easy to predict from a forecast run initialized 12 hours earlier. Thus data preparation is likely one factor in the inability to extract a clear center when the storm is a tropical depression. 
Tropical depressions are not well-organized systems, and whether or not the storm eventually develops a clear eye is highly dependent on the 3D evolution of the storm.  So at this early stage, it is not surprising that a system that uses only a 2D slice of the wind field cannot detect the storm.

\begin{figure}[!t]
    \centering
    \includegraphics[width=\columnwidth]{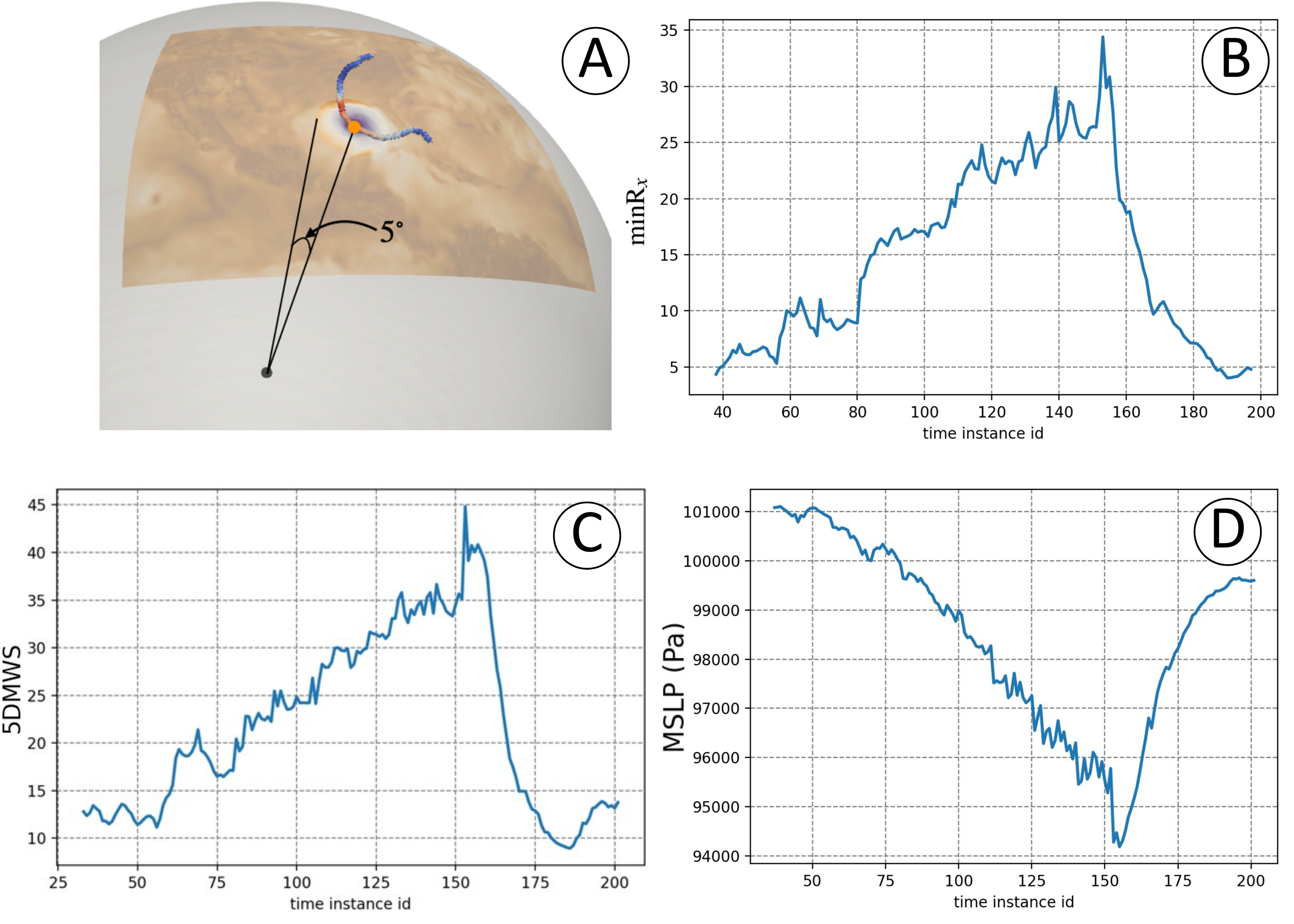}
    \vspace{-6mm}
    \caption{ (A) A schematic in calculating a five degree region. (B) Minimum multilevel robustness $\minR_x$, (C) 5DMWS, and (D) MSLP  along the Katrina trajectory.} 
    \label{fig:windSpeed}
    \vspace{-2mm}
\end{figure}

\begin{figure}[!t]
    \centering
    \includegraphics[width=\columnwidth]{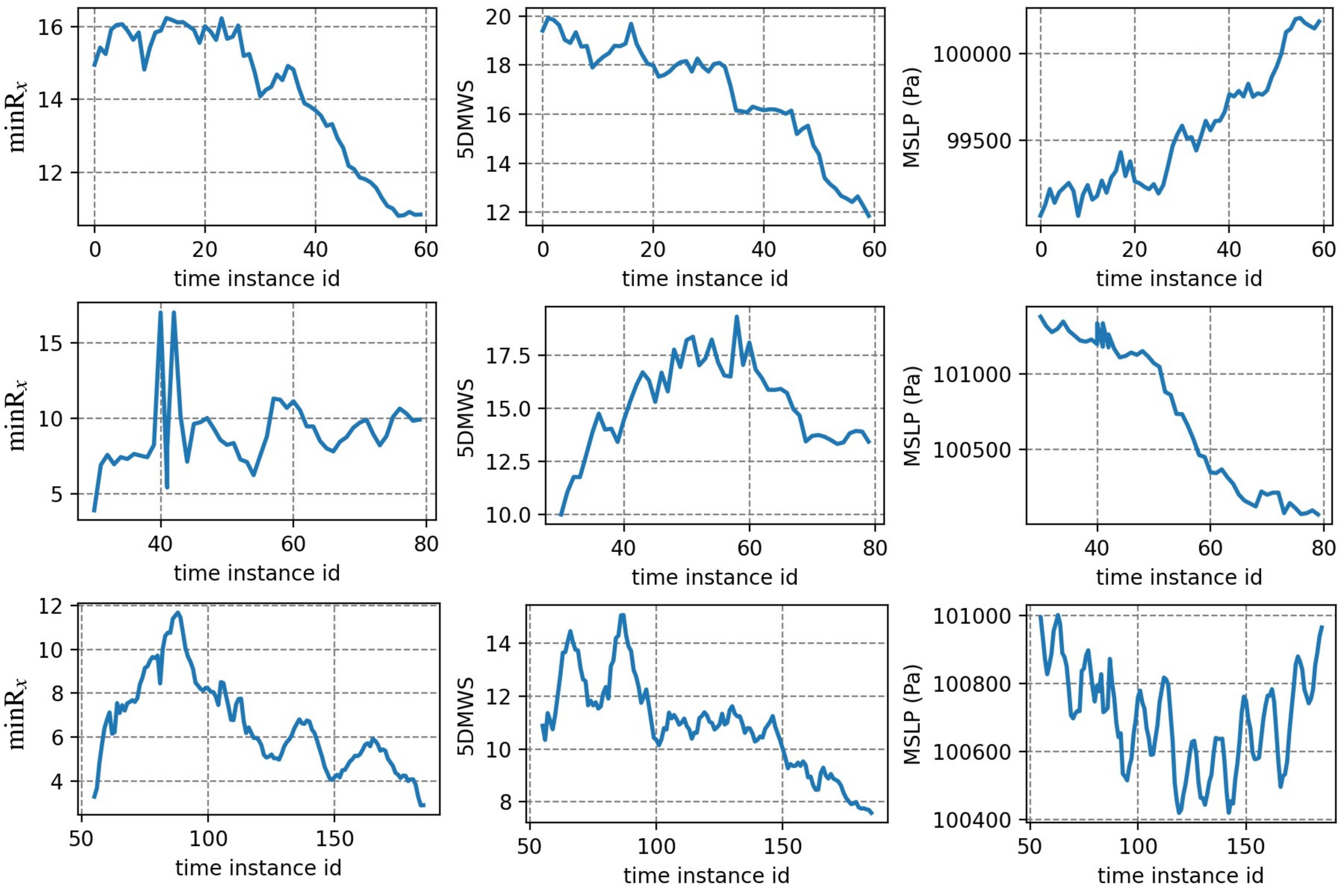}
    \vspace{-6mm}
    \caption{From top to bottom, $\minR_x$, 5DMWS, and MSLP along trajectories $a$, $b$, and $d$ from \autoref{fig:HK}. } 
    \label{fig:quantity-abd}
    \vspace{-2mm}
\end{figure}

\begin{figure*}[t]
    \centering
    \includegraphics[width=1.0\linewidth]{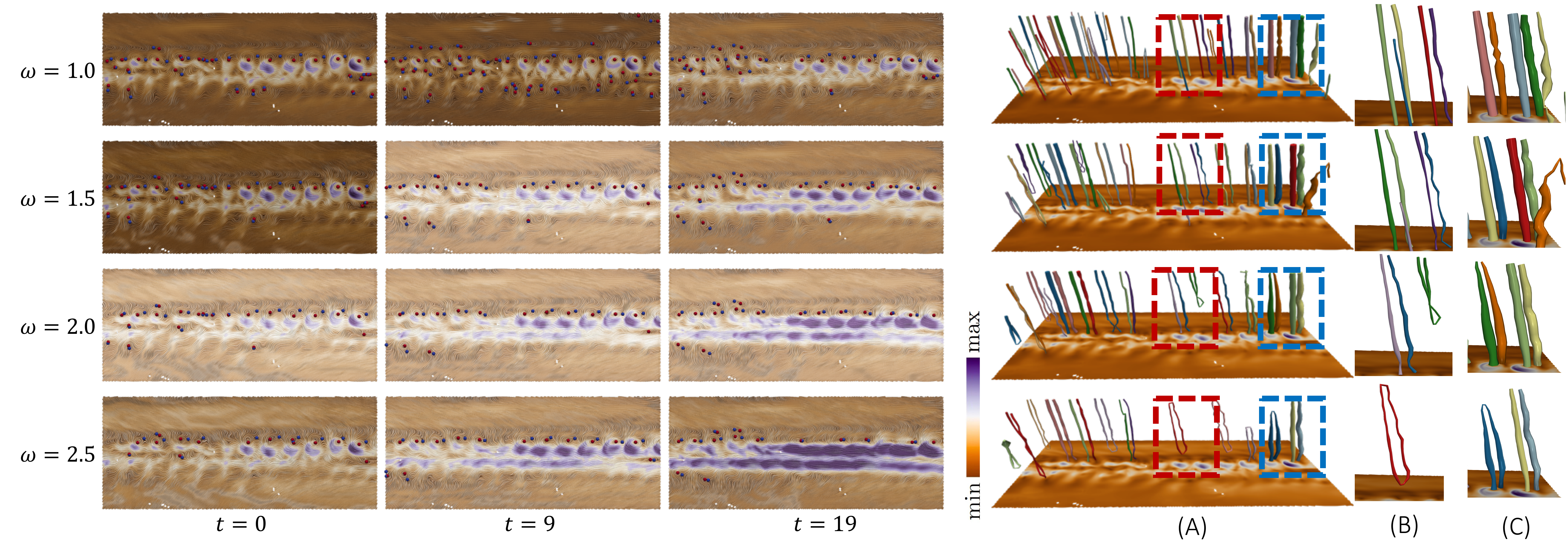}
    \vspace{-6mm}
    \caption{Feature comparison with multilevel robustness for the $\ME$ dataset. Left three columns: selected vector fields at time steps $0, 9$, and $19$, with $\omega =1.0, 1.5, 2.0$, and $2.5$. Critical points with degrees $+1$ and $-1$ are in red and blue, respectively. 
     Columns (A), (B) and (C): feature tracking and selection results and their zoomed-in views; trajectories are colored by their indices with radius proportional to $\minR_x$. (B) corresponds to the regions in (A) surrounded by red boxes, and (C) refers to the regions enclosed by blue boxes.} 
    \label{fig:ME}
    \vspace{-2mm}
\end{figure*}

\para{Correlation with known physical quantities.}
We investigate the relationship between robustness and quantities that are used by the tropical cyclone research community, including five-degree maximum wind speed (5DMWS) and mean sea-level pressure (MSLP).
These two quantities are commonly used by domain scientists to detect, track, and evaluate tropical cyclones. 
Compared with traditional cyclone  tracking schemes, our multilevel robustness framework has the advantage of identifying cyclonic features using only the wind vector fields.

We observe a strong correlation between robustness and 5DMWS, which has been widely used in hurricane intensity metrics such as the Saffir-Simpson scale.  As illustrated in~\autoref{fig:windSpeed}(A),
5DMWS is defined as the maximum wind speed within the five degree region of the hurricane center $x$.  
As shown in~\autoref{fig:windSpeed}(B) and (C), the Pearson correlation coefficients between the $\minR_x$ curve  
and 5DMWS is 0.95, suggesting a strong relationship between these two quantities. 
Mean sea-level pressure (MSLP) is another scalar field that is frequently used by domain scientists in hurricane analysis. MSLP is connected to maximum wind speed via gradient wind balance, modified to account for surface friction; an empirical relationship connecting these two quantities is described in~\cite{Holland2008}. 
Hurricanes higher on the Saffir-Simpson scale (with higher maximum wind speed) have lower MSLP at their center. 
The MSLP along the detected Katrina trajectory is shown in~\autoref{fig:windSpeed}(D).
The Pearson correlation coefficient between the $\minR_x$ curve and MSLP is -0.83. 

We further investigate the correlations between robustness with 5DMWS and MSLP for trajectories $a$, $b$, and $d$ from  \autoref{fig:HK}. 
As shown in \autoref{fig:quantity-abd} (1st row), $\minR_x$ has a strong correlation with 5DMWS and MSLP along trajectory $a$, where the  correlation of the former is 0.94 and of the latter is -0.95.  
Further investigation reveals that trajectory $a$ corresponds to Hurricane Hilary. 
This experiment shows that robustness strongly correlates with physical quantities for trajectories $a$ and $c$, which correspond to Hurricane Hilary (category 2 hurricane) and Hurricane Katrina (category 5 hurricane) respectively.
However, such high correlations do not generalize to weaker storm systems.  
The correlation between $\minR_x$ with 5DMWS is $0.764$ for trajectory $d$, and $-0.032$ for trajectory $b$, see \autoref{fig:quantity-abd} (2nd and 3rd row). 
It turns out that trajectory $d$ represents the Tropical Storm Irwin, whereas trajectory $b$ can not be found in the National Hurricane Center's Tropical Cyclone Reports.

\subsection{Feature Comparison for Ensemble Dataset}
\label{sec:ME}

We demonstrate our framework in feature comparison in an ensemble of global ocean dataset (referred to as $\ME$),  simulated with different wind stress parameters.  The simulation code, MPAS-Ocean~\cite{GolazCaldwellVan2019, PetersenAsayBerres2019}, is a multiscale and unstructred mesh simulation for studying the ocean component of climate changes. 
In this experiment, we utilized the standard low-resolution EC60to30 mesh, whose size of the cells along the coast varied from 60 km to 30 km.
Specifically, in the $\ME$ dataset, each of the four simulation runs captures 20-day ocean eddies with the \emph{bulk wind stress amplification parameter} $\omega$ varying from $1.0$ to $2.5$; the time-resolution of the data is 1 day.  
We truncate the region near the equator in the Pacific Ocean ($15$ S$^\circ$ to $18$ N$^\circ$ for latitude and $170$ E$^\circ$ to $110$ W$^\circ$ for longitude), since this region contains many large eddies for feature comparison. 
~\autoref{fig:ME} (left three columns) visualizes selected vector fields associated with time steps $0$, $9$, and $19$, for $\omega = 1.0, 1.5, 2.0$ and $2.5$, respectively.
In the rest of this section, we investigate the variability of features -- in particular, the centers of eddies -- induced by varying wind stress. 

From the visualization of vector fields in~\autoref{fig:ME}, we obtain some preliminary observations: (1) critical point locations share a similar distribution at the beginning of each simulation (1st column); (2) as $\omega$ increases, vector field features show more variations as $t$ increases (3rd column); (3) a large value of $\omega$ leads to a higher flow magnitude. However, simply showing the locations of critical points in the ensemble has a limited effect on guiding parameter selection and post hoc analysis. 
Instead, our framework can capture more variability across the four parameter settings for feature comparison. 
To preserve the merging and splitting behavior of critical points, we set the threshold for degree filter at $-1$, thereby preserving saddles in the domain. 
We also set $\sigma=0.2$ and the threshold for stability filter at $0.02$ to postprocess initial FTK trajectories and eliminate visual clutter. 

The first observation from our framework is that trajectories have a shorter lifespan as $\omega$ increases; see the trajectories in~\autoref{fig:ME}(B) (from top to bottom) for examples. 
The second observation is that as $\omega$ increases, a number of critical points have decreased robustness and more consistent cancellation partners, see~\autoref{fig:ME}(C). 


The domain scientists pointed out that an increased wind stress will reduce the scales of existing eddies and suppress the development of larger scale eddies. 
This also leads to a decrease in stability, measured by robustness, for some eddy centers as they interact more easily with nearby features, thus locating more consistent cancellation partners (see~\autoref{fig:ME}(C) at $\omega=2.5$). 
As a consequence, some trajectories are filtered out in~\autoref{fig:ME}(A) as $\omega$ increases, since these trajectories become less stable. 
To summarize, our feature comparison captures variability and stability among critical point trajectories under various parameter settings, which may help guide parameter selection (\eg, maintaining a certain number of stable features) in scientific simulations.

%% file: sec-discussions.tex
\section{Conclusion and Discussion}
\label{sec:discussion}

In this paper, we introduce a new multilevel robustness framework.  
Our framework helps to mitigate the drawbacks of the classic robustness computation due to the boundary effect, and better differentiate the behaviors of critical points in terms of their multiscale stability. 
We show that the statistical information of multilevel robustness, in particular, minimum multilevel robustness, can be integrated seamlessly with feature tracking algorithms such as FTK as a postprocessing step.
Our framework thus supports feature tracking, selection, and comparison, and improve the visual interpretability of vector fields from scientific simulations. 

Modern heuristic tracking schemes detect tropical cyclones through a two-step procedure: first, isolated minima in the sea level pressure field are identified; second, an upper-level warm core criteria is used to filter out storms that are not tropical in nature. 
Compared with such heuristics, our robustness-based framework  has the advantage of identifying strong cyclonic features using only the wind vector fields. 

There are a number of points for discussions, such as feature tracking in 3D, scalability, uncertainty visualization, and alternative strategies. 
First, it is possible to extend our current approach to 3D vector fields, since robustness has been studied for critical points in 3D~\cite{SkrabaRosenWang2016}.   
However, topological features such as vortices and vortex cores are arguably more interesting to study in 3D than critical points, where a notion of robustness has yet to be developed. This presents a current limitation of our framework. 
Second, our current implementation approximates multilevel robustness with a discrete set of radii. Increasing the number of levels will require more computational resources, where advanced parallel/distributed computation may be needed (c.f.,~\autoref{fig:runningTime}, which uses an embarrassingly parallel approach). 
Third, our work is motivated by the computation of classic robustness, which may produce artifacts when the sublevel sets containing critical points intersect the domain boundaries. 
It would be interesting to consider alternative strategies. For instance, moving critical points out of the domain by a perturbation may change the structure of the underlying sublevel sets, and the corresponding merge tree may become inconsistent with the original (observable) data.
Fourth, the multilevel robustness is a natural candidate for uncertainty visualization, which is left for future work. 
Finally, the classic robustness has been applied to data beyond climate science, such as vector fields from combustion simulation and tensor fields from materials science and diffusion tensor imaging. We believe a generalization of multilevel robustness to these datasets would be interesting but beyond the scope of the current paper. A main challenge is to study its correlation with physical quantities in these respective application domains.

%% file: ml-robustness-refs.bbl
\newcommand{\etalchar}[1]{$^{#1}$}